\documentclass{article}
\usepackage{arxiv}

\usepackage{times}

\usepackage{soul}
\usepackage{url}
\usepackage[utf8]{inputenc}
\usepackage[small]{caption}
\usepackage{graphicx}
\usepackage{amsmath}
\usepackage{booktabs}
\usepackage{algorithm}
\usepackage{algorithmic}
\usepackage{bm}
\usepackage{amssymb}
\usepackage{amsfonts}
\usepackage{xcolor}
\usepackage{amsthm}
\usepackage{subcaption}
\usepackage{multirow}
\usepackage{comment}
\usepackage{dcolumn}
\theoremstyle{definition}
\newtheorem{assumption}{Assumption}
\theoremstyle{plain}
\newtheorem{theorem}{Theorem}
\newtheorem{corollary}{Corollary}
\newcommand\independent{\protect\mathpalette{\protect\independenT}{\perp}}
\def\independenT#1#2{\mathrel{\rlap{$#1#2$}\mkern2mu{#1#2}}}

\title{Estimation of Local Average Treatment Effect by Data Combination}

\author{
Kazuhiko Shinoda\\
Graduate School of Economics, Keio University\\
\And
Takahiro Hoshino\\
Faculty of Economics, Keio University\\
RIKEN AIP
}
\date{\empty}

\begin{document}

\maketitle

\begin{abstract}
It is important to estimate the local average treatment effect (LATE) when compliance with a treatment assignment is incomplete.
The previously proposed methods for LATE estimation required all relevant variables to be jointly observed in a single dataset; however, it is sometimes difficult or even impossible to collect such data in many real-world problems for technical or privacy reasons.
We consider a novel problem setting in which LATE, as a function of covariates, is nonparametrically identified from the combination of separately observed datasets.
For estimation, we show that the direct least squares method, which was originally developed for estimating the average treatment effect under complete compliance, is applicable to our setting.
However, model selection and hyperparameter tuning for the direct least squares estimator can be unstable in practice since it is defined as a solution to the minimax problem.
We then propose a weighted least squares estimator that enables simpler model selection by avoiding the minimax objective formulation.
Unlike the inverse probability weighted (IPW) estimator, the proposed estimator directly uses the pre-estimated weight without inversion, avoiding the problems caused by the IPW methods.
We demonstrate the effectiveness of our method through experiments using synthetic and real-world datasets.
\end{abstract}

\section{Introduction}
Estimating the causal effects of treatment on an outcome of interest is central to optimal decision making in many real-world problems, such as policymaking, epidemiology, education, and marketing \cite{skovron2015practical,wood2008empirical,oreopoulos2006estimating,varian2016causal}.
However, the identification and estimation of treatment effects usually relies on the untestable assumption referred to as \emph{unconfoundedness}, namely, independence between the treatment status and potential outcomes \cite{imbens2015causal}.
Violations of unconfoundedness may occur not only in observational studies, but also in randomized controlled trials (RCTs) when compliance to the assigned treatment is not complete.
For example, even if a coupon is distributed randomly to measure its effect on sales, the probability of using the coupon is likely to depend on the unobserved nature of the individuals.
Moreover, noncompliance can also occur regardless of the individual's intentions.
In online advertisement placement, the probability of watching the ad depends on the bidding strategy of other companies because ads that are actually displayed are determined through the real-time-bidding even if one tries to randomly place the ad.

In such cases, it is well known that \emph{the local average treatment effect (LATE)} can be identified and estimated using the treatment assignment as an instrumental variable (IV) and conditions milder than unconfoundedness \cite{Imbens1994-mm,Angrist1996-cb,Frolich2007-hc}.
LATE is the treatment effect measured for the subpopulation of compliers, individuals who always follow the given assignment.

We suppose that we cannot observe all relevant variables in a single dataset for technical or privacy reasons.
For example, in online-to-offline (O2O) marketing, where treatments are implemented online and outcomes are observed offline, it is often difficult to match the records of the same individuals observed separately online and offline.
Additionally, with the global anti-tracking movement gaining momentum, it may become more difficult to combine multiple pieces of information online as well.
Although causal inference by data combination has been actively studied \cite{Ridder2007-an,Bareinboim2016-vk,Lee_Correa_Bareinboim_2020}, LATE estimation using multiple datasets has not received much attention despite its practical importance.
We extend the problem setting considered in \cite{yamane}, where two different treatment regimes are available, to allow for the existence of noncompliance.

For the estimation, we show that the direct estimation method originally developed for the average treatment effect (ATE) under the complete compliance \cite{yamane} can be applied to the LATE estimation in our setting.
However, their method has a practical issue in that model selection and hyperparameter tuning can be unstable owing to its minimax objective formulation.
We then propose a weighted least squares estimator to avoid the minimax objective, and improve the stability in practice.
Unlike the inverse probability weighted (IPW) estimator \cite{wooldridge2002,Wooldridge2007-vn,Seaman2013-mw}, which is often employed to estimate treatment effects, the proposed estimator directly uses the estimated propensity-score-difference (PSD) as a weight without inversion.
Therefore, our method can also avoid the common issue in the IPW methods, that is, high variance at points with a propensity score extremely close to zero.

The contributions of this study lie in the following three parts.
First, we show that LATE is identified even when an outcome and treatment status cannot be observed simultaneously in a single dataset, and the treatment assignment is completely missing.
Second, we find that the positivity assumption, which is necessary in the standard setting with one regime, can be omitted in our setting.
We show this relaxation of the conditions further facilitates data collection.
Third, we develop a novel estimator that enables simpler model selection while maintaining the essence of direct estimation as much as possible.

\section{Problem Setting}
Unlike the standard causal inference studies, we consider a setting where there are two different assignment regimes \cite{yamane}, which we term as \emph{the two-regime design (2RD)}.
The concept is quiet general that it only requires two observational studies, two RCTs or a combination of the two.
We have to be assured that the treatment assignment is done based on different regimes, i.e. different probabilities (see Assumption \ref{as:regime}.\ref{as:2asgn}).

We define our problem using the potential outcome framework \cite{rubin1974,imbens2015causal}.
Let $K\in\{0,1\}$ be a regime indicator, and we use a superscript $k=0,1$ to specify the regime which the variables come from.
Let $Y^{(k)}\in\mathcal{Y}\subset\mathbb{R}$ be an outcome of interest, $D^{(k)}\in\{0,1\}$ be a binary treatment indicator, $Z^{(k)}\in\{0,1\}$ be an assignment indicator and $\bm{X}^{(k)}\in\mathcal{X}\subset\mathbb{R}^{q_x}$ be a $q_x$-dimensional vector of covariates.
$D_z^{(k)}$ is the potential treatment status realized only when $Z^{(k)}=z$, and $Y_d^{(k)}$ is the potential outcome realized only when $D^{(k)}=d$, where $z,d\in\{0,1\}$.
Using this notation, we implicitly assume that $Z^{(k)}$ does not have a direct effect on $Y^{(k)}$, but affects $Y^{(k)}$ indirectly through $D^{(k)}$.
This condition, often referred to as the exclusion restriction, is necessary for $Z^{(k)}$ to be a valid IV.
Let $Y_1$, $Y_0$, $D_1$, $D_0$ and $\bm{X}$ be the potential variables and covariates in the population of interest, and we suppose that the iid samples from $P(\bm{X})$ can be obtained as test samples.

\begin{figure}[t]
    \begin{center}
    \includegraphics[width=0.7\textwidth]{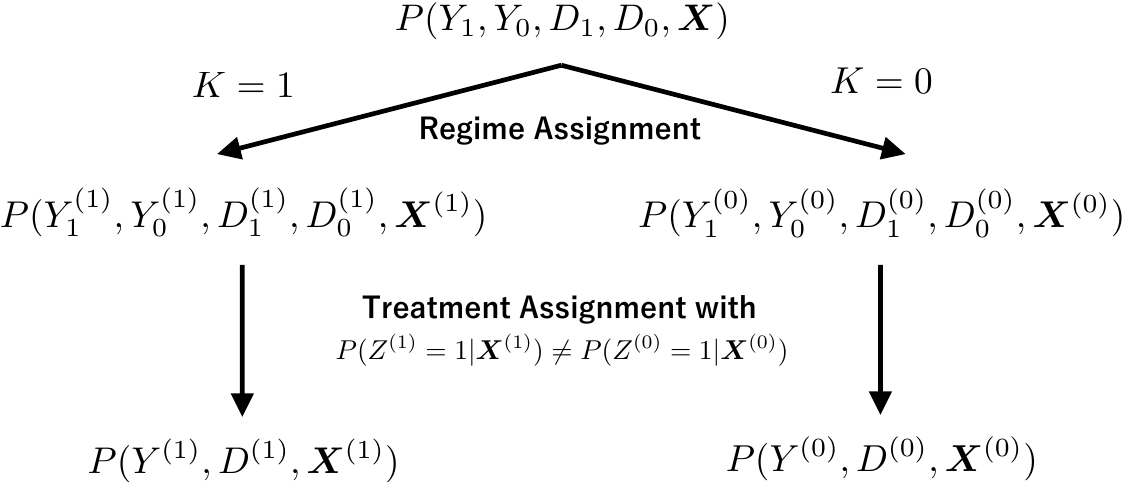}
    \caption{Overview of the 2RD.}
    \label{fig:2RD}
    \end{center}
\end{figure}

\subsection{Basic Setting}
First, we make the following assumption on the relation between the two regimes.
\begin{assumption}\label{as:regime}
\!
\begin{enumerate}
\item $P(Y_1^{(1)},Y_0^{(1)},D_1^{(1)},D_0^{(1)},\bm{X}^{(1)})=P(Y_1^{(0)},Y_0^{(0)},D_1^{(0)},D_0^{(0)},\bm{X}^{(0)})=P(Y_1,Y_0,D_1,D_0,\bm{X})$.\label{as:invar}
\item $P(Z^{(1)}=1|\bm{X}^{(1)}=\bm{x})\neq P(Z^{(0)}=1|\bm{X}^{(0)}=\bm{x})$ for any $\bm{x}\in\mathcal{X}$.\label{as:2asgn}
\end{enumerate}
\end{assumption}
By Assumption \ref{as:regime}.\ref{as:invar}, we suppose that the joint distribution of the potential variables and covariates in each assignment regime is invariant to each other, and equal to the joint distribution in the population of interest.
This means that the participants in each regime are random draws from the population of interest.
We can still identify LATE even if we weaken Assumption \ref{as:regime}.\ref{as:invar} to its conditional version, but the direct estimation is no longer possible.
See Appendix A for more discussion.
Assumption \ref{as:regime}.\ref{as:2asgn} indicates that the assignment regimes are different to each other for any level of the covariates.

Additionally, we make the following assumption required for the identification of LATE.
Here, $A\independent B|C$ means $A$ and $B$ are conditionally independent given $C$.
\begin{assumption}\label{as:standard}
For $k=0,1$,
\begin{enumerate}
\item $Y_1^{(k)},Y_0^{(k)},D_1^{(k)},D_0^{(k)}\independent Z^{(k)}|\bm{X}^{(k)}$.\label{as:uncon}
\item $D^{(k)}=Z^{(k)}D_1^{(k)}+(1-Z^{(k)})D_0^{(k)}$ and $Y^{(k)}=D^{(k)}Y_1^{(k)}+(1-D^{(k)})Y_0^{(k)}$.\label{as:consis}
\item $P(D_1^{(k)}\geq D_0^{(k)}|\bm{X}^{(k)})=1$.\label{as:mono}
\item $P(D_1^{(k)}=1|\bm{X}^{(k)})\neq P(D_0^{(k)}=1|\bm{X}^{(k)})$.\label{as:inst}
\end{enumerate}
\end{assumption}
Assumption \ref{as:standard}.\ref{as:uncon} states that $Z^{(k)}$ are randomly assigned within a subpopulation sharing the same level of the covariates.
The mean independence between the potential variables and the treatment assignment conditional on covariates is sufficient for identifying LATE, but we maintain Assumption \ref{as:standard}.\ref{as:uncon} as it is typical to assume the full conditional independence in the causal inference literature.
Assumption \ref{as:standard}.\ref{as:consis} relates the potential variables to their realized counterparts.
Assumption \ref{as:standard}.\ref{as:mono} excludes defiers, those who never follow the given treatment assignment, from our analysis.
Assumption \ref{as:standard}.\ref{as:inst} is necessary for $Z^{(k)}$ to be a valid IV of $D^{(k)}$.
Figure \ref{fig:2RD} illustrates the data generating process under the 2RD.

Assumption \ref{as:standard} is the direct extension of the standard assumptions used for the LATE estimation \cite{abadie2003,Frolich2007-hc,Tan2006-el,ogburn2015} to the 2RD.
However, we omit the condition referred to as positivity $0<P(Z^{(k)}=1|\bm{X}^{(k)})<1$ as it is unnecessary in the 2RD.
Therefore, it is possible to set $P(Z^{(0)}=1|\bm{X}^{(0)})=0$ as long as Assumption \ref{as:regime}.\ref{as:2asgn} is satisfied.
We term the design with $P(Z^{(1)}=1|\bm{X}^{(1)})>0$ for any $\bm{X}^{(1)}$ and $P(Z^{(0)}=1|\bm{X}^{(0)})=0$ for any $\bm{X}^{(0)}$ as \emph{the one-experiment 2RD (1E2RD)} since an experiment is conducted only for those with $k=1$, and a natural state without any intervention is observed for those with $k=0$.

Although \cite{yamane} does not mention anything on this point, it is of great practical importance since the 1E2RD is much easier to implement than the general 2RD, and extends the plausibility of our setting.
In the 1E2RD, we only have to conduct one experiment or collect one set of observational data just like the standard setting for causal inference.
All we need in addition is a dataset collected from those without any intervention.
Such data may be available at no additional cost, for example, when there is appropriate open data published by the government.

Hereafter, we abuse the notation by omitting the superscript on the potential variables and covariates unless this causes confusion.

\subsection{Identification}
The parameter of our interest is LATE as a function of covariates $\bm{X}$, defined as
\begin{align*}
\mu(\bm{X}):=E[Y_1-Y_0|D_1>D_0,\bm{X}].
\end{align*}
It measures how covariates $\bm{X}$ affect the average treatment effect within the subpopulation of compliers.
The following theorem shows that $\mu(\bm{X})$ is nonparametrically identified in an unusual form in our setting.
\begin{theorem}\label{th:identify}
Under Assumption \ref{as:regime} and \ref{as:standard},
\begin{align}
\mu(\bm{X})=\frac{E[Y^{(1)}|\bm{X}]-E[Y^{(0)}|\bm{X}]}{E[D^{(1)}|\bm{X}]-E[D^{(0)}|\bm{X}]}.\label{eq:ident}
\end{align}
\end{theorem}
The proof can be found in Appendix B.
Notably, this form coincides with the identification result of ATE in \cite{yamane}, which means that we can also estimate $\mu(\bm{X})$ from the same combination of datasets proposed in \cite{yamane} under appropriate assumptions.
This is a rather powerful result in practice since $Z^{(k)}$ is not required despite the presence of noncompliance.
Moreover, it is obvious that $p_d^{(k)}:=P(D^{(k)}=1)$ and $P(\bm{X}|D^{(k)}=1)$ are sufficient to identify the denominator of (\ref{eq:ident}) since $E[D^{(k)}|\bm{X}]$ can be identified as $\frac{P(\bm{X}|D^{(k)}=1)p_d^{(k)}}{P(\bm{X})}$ by applying the Bayes' theorem.

The point is that we can identify $\mu(\bm{X})$ as a combination of functions depending only on covariates $\bm{X}$ in our setting.
This property allows us to use a direct estimation method for $\mu(\bm{X})$.
We can identify LATE as $\mu(\bm{X})=\frac{E[Y|Z=1,\bm{X}]-E[Y|Z=0,\bm{X}]}{E[D|Z=1,\bm{X}]-E[D|Z=0,\bm{X}]}$ under the standard assumptions \cite{abadie2003,Frolich2007-hc,Tan2006-el,ogburn2015}, but we cannot rewrite it as a combination of functions depending only on covariates (see Appendix A).

Theorem \ref{th:identify} also suggests the usefulness of the 1E2RD.
We can simplify (\ref{eq:ident}) when we implement the 1E2RD under \emph{one-sided noncompliance}, where individuals assigned to the control group never receive treatment, but they have a choice if assigned to the treatment group.
\begin{corollary}
Assume $P(Z^{(0)}=1|\bm{X})=0$ (1E2RD) and $P(D_0=1|\bm{X})=0$ (one-sided noncompliance) in addition to Assumption \ref{as:regime} and \ref{as:standard}.
Then, $E[Y^{(0)}|\bm{X}]=E[Y_0|\bm{X}]$ and
\begin{align*}
\mu(\bm{X})=\frac{E[Y^{(1)}|\bm{X}]-E[Y^{(0)}|\bm{X}]}{E[D^{(1)}|\bm{X}]}.
\end{align*}
\end{corollary}
This corollary shows that we can reduce the number of necessary datasets in the 1E2RD under one-sided noncompliance.
This fact not only facilitates the data collection, but also benefits the estimation since the denominator is now just a propensity score, thus estimable accurately by, for example, the Positive-Unlabeled (PU) learning \cite{elkannoto2008,duPlessis2014,duPlessis2015,Kiryo2017-pz} with the logistic loss.

Although one-sided noncompliance is often associated with RCTs, some observational studies also fit with the framework \cite{Frolich2013-sk,Kennedy2020-wz}.
In the case of one-sided noncompliance, we can also consider our problem as the estimation of the average treatment effect on the treated (ATT) since LATE is equal to ATT under one-sided noncompliance and the other standard assumptions \cite{Frolich2013-sk,Donald2014-ca}.

\subsection{Data Collection Scheme}
We assume that the joint samples of $(Y^{(k)},D^{(k)},Z^{(k)},\bm{X})$ are not available.
By Theorem \ref{th:identify}, the following separate datasets and the estimate of $p_d^{(k)}$ are sufficient for estimating $\mu(\bm{X})$:
\begin{align*}
\{\bm{x}_{di}^{(k)}\}_{i=1}^{n_d^{(k)}}\overset{\mathrm{iid}}{\sim}P(\bm{X}|D^{(k)}=1),\hspace{5mm}\{(y_i^{(k)},\bm{x}_i^{(k)})\}_{i=1}^{n^{(k)}}\overset{\mathrm{iid}}{\sim}P(Y^{(k)},\bm{X}),
\end{align*}
for $k=0,1$.
This setting is much easier to apply to real-world situations than the standard setting where the joint samples are required for every individual.

We can interpret this data collection scheme in two ways.
First, it can be regarded as a version of \emph{the repeated cross-sectional (RCS) design} \cite{Moretti2004-nt,Athey2006,Guell2006-wu,Ridder2007-an,lebo2015effective} with $k=0,1$ representing the time points before and after the assignment regime switches, respectively.
The 1E2RD is also possible in this case by setting $P(Z^{(0)}=1|\bm{X}^{(0)})=0$.
This means collecting data for $k=0$ at some point before an experiment is conducted.
Generally, although panel data are advantageous for statistical analyses because they follow the same individuals over multiple time periods, RCS data have some advantages over panel data.
RCS data are easier and cheaper to collect in many cases.
Consequently, they are often more representative of the population of interest and larger in sample size than panel data.

However, there is a concern regarding the validity of Assumption \ref{as:regime}.\ref{as:invar} when we use RCS data since the potential variables may change over time.
We need some sort of side information to be confident about the use of RCS data in our setting as we cannot directly test whether the potential variables remain unchanged.

The second interpretation is to collect data with $k=0,1$ at the same time by randomly splitting the population of interest.
If an experimenter is able to surely perform random splitting, this approach is favorable since we do not have to worry about the validity of Assumption \ref{as:regime}.\ref{as:invar}.
The implementation of random splitting is easy in the case of RCTs.

One specific example of our setting is O2O marketing.
To measure the effect of an online video ads on sales in physical stores, we usually cannot link the data of those who watched the ads with that of those who shopped at the stores.
However, it is relatively easy to separately collect data from those who watch the ads and the purchasers.
In this case, data collection can be performed by either RCS design or random splitting.

Another example is when estimating the effect of a treatment that takes a certain period of time to take effect.
For instance, it is desirable to use panel data to estimate the effect of job training on future earnings.
However, individuals may gradually drop out of the survey, and the probability of the attrition can depend on unobserved variables \cite{Hirano2001-hm,Nevo2003-es}.
Therefore, it is easier to collect RCS data than to construct balanced panel data.
Because we do not need to observe the outcome and treatment status simultaneously, we are more likely to be able to collect even larger and more representative data than with a normal RCS design.

\section{Related Works}
There have been many proposals for the estimation of $\mu(\bm{X})$ from the joint samples \cite{Little1998-fk,Hirano2000-uy,abadie2003,Tan2006-el,Okui2012-wy,ogburn2015,Wang2021-xk}.
These include estimation via the parametric specification of the local average response function $E[Y_d|D_1>D_0,\bm{X}]$ \cite{abadie2003}, doubly robust estimators \cite{Okui2012-wy,ogburn2015} and estimation with a binary outcome \cite{Wang2021-xk}, to name a few.
However, the estimation of $\mu(\bm{X})$ by data combination has rarely been considered in the literature.

The 2RD has a similar structure to that of \emph{the instrumented difference-in-differences (IDID)} \cite{Ye2020InstrumentedD}.
The main differences between them are: the conditional independence assumption in IDID is strictly milder than our Assumption \ref{as:standard}.\ref{as:uncon}, but IDID requires $Z$, and the direct estimation is not possible.
Which setting is more plausible and practical depends on an actual situation.

Our problem setting is also closely related to \emph{the two-sample IV (TSIV)} estimation \cite{angrist1992,inoue2010two,pacini2016robust,choi,buchinsky2021estimation}, where the outcome, IVs, and covariates are not jointly observed in a single dataset.
Furthermore, the idea of using moments separately estimated from different datasets can be dated back to at least \cite{klevmarken1982missing}.
Although estimands in the TSIV estimation are not limited to causal parameters, there have been some studies on the causal inference in the setting related to TSIV.
They include ATE estimation from samples of $(Y,Z,\bm{X})$ and $(D,Z,\bm{X})$ with the existence of unmeasured confounders \cite{Sun_2022} and causal inference using samples from heterogeneous populations \cite{Zhao2019-xl,shutan}.
Our setting differs from theirs in that we have to observe $D$ only when $D=1$, and we do not need $Z$ at all.
Particularly, it is of great practical benefit since samples of those who do not receive treatment are often rather difficult to observe.
Although our setting requires two regimes, it rather opens up the possibility of LATE estimation in the real world since the 1E2RD is possible.

Another approach for the causal inference from separately observed samples is \emph{the partial identification} \cite{manski2003partial,tamer2010partial,molinari2020microeconometrics}, that is, deriving bounds for the treatment effects rather than imposing strong assumptions sufficient for the point identification.
In \cite{Fan2014-ir,Fan2016-gp}, the moment inequality proposed in \cite{cambanis1976inequalities} was applied to derive the sharp bounds for ATE when only samples of $(Y,D)$ and $(D,\bm{X})$ are available.
The difficulty of applying their approach to the estimation of a treatment effect conditional on covariates is that it requires conditional distribution functions or conditional quantile functions, which are usually difficult to estimate accurately.
Moreover, it is sometimes difficult to derive informative bounds for treatment effects without imposing strong assumptions depending on the data generating process \cite{molinari2020microeconometrics}.

\section{Existing Methods}\label{sec:exist}
Although the LATE estimation in our specific setting has not been studied before, some existing methods can be applied.
We discuss the advantages and disadvantages of these methods especially in terms of accuracy and model selection.
Since the true value of treatment effects is not observable by \emph{the fundamental problem of causal inference} \cite{holland1986statistics}, model selection and hyperparameter tuning are substantial issues in practice \cite{rolling2014model,alaa2018limits,saito2020}.

\subsection{Separate Estimation}
A naive estimation method for $\mu(\bm{X})$ is to separately estimate the components and combine them as
\begin{align}
\widehat{\mu}_{\mathrm{sep}}(\bm{x})=\frac{\widehat{E}[Y^{(1)}|\bm{X}=\bm{x}]-\widehat{E}[Y^{(0)}|\bm{X}=\bm{x}]}{\widehat{E}[D^{(1)}|\bm{X}=\bm{x}]-\widehat{E}[D^{(0)}|\bm{X}=\bm{x}]},\label{eq:sep}
\end{align}
where a hat denotes an estimator.
$E[D^{(k)}|\bm{X}]$ can be estimated by the PU learning \cite{elkannoto2008,duPlessis2014,duPlessis2015,Kiryo2017-pz} with the logistic loss by using $\{\bm{x}_{di}^{(k)}\}_{i=1}^{n_d^{(k)}}$ as positive data, and the covariates in $\{(y_i^{(1)},\bm{x}_i^{(1)})\}_{i=1}^{n^{(1)}}$ and $\{(y_i^{(0)},\bm{x}_i^{(0)})\}_{i=1}^{n^{(0)}}$ as unlabeled data.
Separate estimation is easy to implement, but usually does not provide a good estimate since it has four possible sources of error.

One advantage of the separate estimation is that the model selection can also be naively performed by choosing the best model for each component.
We can easily calculate the model selection criteria, such as the mean squared error (MSE) for each component from the separately observed datasets.
However, the resulting $\widehat{\mu}_{\mathrm{sep}}$ may perform poorly because it does not necessarily minimize the model selection criterion in terms of the true $\mu$ \cite{rolling2014model}.

We can alleviate the drawback of the separate estimation by directly estimating the numerator and denominator in (\ref{eq:sep}).
Let $T:=KD^{(1)}-(1-K)D^{(0)}$ and $U:=KY^{(1)}-(1-K)Y^{(0)}$ be the auxiliary variables for the notational and computational convenience, and assume $P(K=1|\bm{X})=0.5$.
Then, we can rewrite the expression in Theorem \ref{th:identify} as $\mu(\bm{X})=\nu(\bm{X})/\pi(\bm{X})$, where $\nu(\bm{X}):=E[U|\bm{X}]$ and $\pi(\bm{X}):=E[T|\bm{X}]$.
To estimate $\nu(\bm{X})$ and $\pi(\bm{X})$, we can construct combined datasets
\begin{align*}
\{(t_i,\bm{x}_{ti},r_{ti})\}_{i=1}^{n_t}&:=\left\{\left((-1)^{1-k},\bm{x}_{di}^{(k)},\frac{\widehat{p}_d^{(k)}(n_d^{(1)}+n_d^{(0)})}{2n_d^{(k)}}\right)\right\}_{i=1,k=0}^{n_d^{(k)},1},\\
\{(u_i,\bm{x}_{ui},r_{ui})\}_{i=1}^{n_u}&:=\left\{\left((-1)^{1-k}y_i^{(k)},\bm{x}_i^{(k)},\frac{n^{(1)}+n^{(0)}}{2n^{(k)}}\right)\right\}_{i=1,k=0}^{n^{(k)},1}
\end{align*}
from $\{\bm{x}_{di}^{(k)}\}_{i=1}^{n_d^{(k)}}$ and $\{(y_i^{(k)},\bm{x}_i^{(k)})\}_{i=1}^{n^{(k)}}$, respectively, where $n_t:=n_d^{(1)}+n_d^{(0)}$ and $n_u:=n^{(1)}+n^{(0)}$.

We can approximate the expectation of a product of any function of $\bm{X}$ and $T$ or $U$ by the simple sample average using the above datasets since we have
\begin{align*}
\frac{1}{n_t}\sum_{i=1}^{n_t}r_{ti}t_if(\bm{x}_{ti})&=\frac{\widehat{p}_d^{(1)}}{2n_d^{(1)}}\sum_{i=1}^{n_d^{(1)}}f(\bm{x}_{di}^{(1)})-\frac{\widehat{p}_d^{(0)}}{2n_d^{(0)}}\sum_{i=1}^{n_d^{(0)}}f(\bm{x}_{di}^{(0)})\approx\frac{1}{2}E[D^{(1)}f(\bm{X})]-\frac{1}{2}E[D^{(0)}f(\bm{X})],\\
\frac{1}{n_u}\sum_{i=1}^{n_u}r_{ui}u_if(\bm{x}_i)&=\frac{1}{2n^{(1)}}\sum_{i=1}^{n^{(1)}}y_if(\bm{x}_i^{(1)})-\frac{1}{2n^{(0)}}\sum_{i=1}^{n^{(0)}}y_if(\bm{x}_i^{(0)})\approx\frac{1}{2}E[Y^{(1)}f(\bm{X})]-\frac{1}{2}E[Y^{(0)}f(\bm{X})].
\end{align*}
An estimator of $\nu(\bm{X})$ can be obtained by any regression method using the above combined dataset, while we need a little twist to obtain an estimator of the PSD $\pi(\bm{X})$.
See Appendix C for the direct estimation methods for the PSD.

\subsection{Direct Least Squares Estimation}\label{sec:dlse}
We can apply \emph{the direct least squares (DLS)} method \cite{yamane} originally proposed for ATE estimation under complete compliance.
Motivated by the drawbacks of the separate estimation, the DLS directly estimates $\mu(\bm{X})$, which is advantageous not only in performance but also in computational efficiency.

The following theorem corresponds to Theorem 1 in \cite{yamane}.
\begin{theorem}\label{th:estlate}
Assume $\nu\in L^2$, where $L^2:=\left\{f:\mathcal{\bm{X}}\mapsto\mathbb{R}\left|E[f(\bm{X})^2]<\infty\right.\right\}$ in addition to Assumption \ref{as:regime} and \ref{as:standard}.
Furthermore, define $H_f(\bm{X}):=\pi(\bm{X})f(\bm{X})-\nu(\bm{X})$.
Then, $\mu$ is equal to the solution of the following least squares problem:
\begin{align}\label{eq:obj_original}
\min_{f\in L^2}E\left[H_f(\bm{X})^2\right].
\end{align}
\end{theorem}
Theorem \ref{th:estlate} immediately follows from Theorem \ref{th:identify}.
In what follows, we show that the minimizer of the problem (\ref{eq:obj_original}) can be estimated without going through the estimation of the conditional mean functions $\nu$ and $\pi$.
Since $\left(H_f(\bm{X})-g(\bm{X})\right)^2\geq0$ for any square integrable function $g\in L^2$, we have $H_f(\bm{X})^2\geq2H_f(\bm{X})g(\bm{X})-g(\bm{X})^2$ by expanding the square.
The equality holds at $g(\bm{X})=H_f(\bm{X})$ for any $f$, which maximizes $2H_f(\bm{X})g(\bm{X})-g(\bm{X})^2$ with respect to $g$.
Hence,
\begin{align}
\mu(\bm{X})=\underset{f\in L^2}{\mathrm{argmin}}\max_{g\in L^2}J(f,g),\label{eq:minimax}
\end{align}
where $J(f,g):=E\left[2H_f(\bm{X})g(\bm{X})-g(\bm{X})^2\right]$.
Rewriting $J(f,g)$ yields
\begin{align}
J(f,g)=2E[Tf(\bm{X})g(\bm{X})]-2E[Ug(\bm{X})]-E[g(\bm{X})^2].\label{eq:obj_dls}
\end{align}
While the objective functional in (\ref{eq:obj_original}) requires the conditional mean estimators, this form can be estimated based on the sample averages as
\begin{align}
\widehat{J}(f,g)=&\frac{2}{n_t}\sum_{i=1}^{n_t}r_{ti}t_if(\bm{x}_{ti})g(\bm{x}_{ti})-\frac{2}{n_u}\sum_{i=1}^{n_u}r_{ui}u_ig(\bm{x}_{ui})-\frac{1}{n_u}\sum_{i=1}^{n_u}r_{ui}g(\bm{x}_{ui})^2.\label{eq:dls_obj_emp}
\end{align}
We can implement the DLS estimation of $\mu$ with an arbitrary regularization term $\Omega$ in practice:
\begin{align*}
\widehat{\mu}_{\mathrm{dls}}(\bm{x})=\underset{f\in F}{\mathrm{argmin}}\max_{g\in G}\widehat{J}(f,g)+\Omega(f,g),
\end{align*}
where $F$ and $G$ are model classes for $f$ and $g$, respectively.

Although any model can be trained by optimizing the model parameters to minimize the above objective functional, a practically useful choice is a linear-in-parameter model.
Set $F=\{f_{\bm{\alpha}}:\bm{x}\mapsto\bm{\alpha}^\top\bm{\phi}(\bm{x})|\bm{\alpha}\in\mathbb{R}^{q_f}\}$ and $G=\{g_{\bm{\beta}}:\bm{x}\mapsto\bm{\beta}^\top\bm{\psi}(\bm{x})|\bm{\beta}\in\mathbb{R}^{q_g}\}$, where $\bm{\phi}$ and $\bm{\psi}$ are $q_f$- and $q_g$-dimensional basis functions, respectively.
Using the $\ell_2$-regularizer, we have
\begin{align*}
\widehat{J}(f,g)+\Omega(f,g)=2\bm{\alpha}^\top\bm{A}\bm{\beta}&-2\bm{\beta}^\top\bm{b}-\bm{\beta}^\top\bm{C}\bm{\beta}+\lambda_f\bm{\alpha}^\top\bm{\alpha}+\lambda_g\bm{\beta}^\top\bm{\beta},
\end{align*}
where
\begin{align*}
\bm{A}:=\frac{1}{n_t}\sum_{i=1}^{n_t}r_{ti}t_i\bm{\phi}(\bm{x}_{ti})\bm{\psi}(\bm{x}_{ti})^\top,\hspace{5mm}\bm{b}:=\frac{1}{n_u}\sum_{i=1}^{n_u}r_{ui}u_i\bm{\psi}(\bm{x}_{ui}),\hspace{5mm}\bm{C}:=\frac{1}{n_u}\sum_{i=1}^{n_u}r_{ui}\bm{\psi}(\bm{x}_{ui})\bm{\psi}(\bm{x}_{ui})^\top,
\end{align*}
and $\lambda_f$ and $\lambda_g$ are some positive constants.
The advantage of this formulation with the linear-in-parameter models and $\ell_2$-regularizer is that we have an analytical solution.
The solution to the inner maximization is given by
\begin{align*}
\widehat{\bm{\beta}}=(\bm{C}+\lambda_g\bm{I}_{q_g})^{-1}(\bm{A}^\top\bm{\alpha}-\bm{b}),
\end{align*}
where $\bm{I}_{q_g}$ denotes the $q_g\times q_g$ identity matrix.
We can then obtain the DLS estimator of $\bm{\alpha}$ by substituting $\widehat{\bm{\beta}}$ and solving the outer minimization:
\begin{align*}
\widehat{\bm{\alpha}}_{\mathrm{dls}}=\left\{\bm{A}(\bm{C}+\lambda_g\bm{I}_{q_g})^{-1}\bm{A}^\top+\lambda_f\bm{I}_{q_f}\right\}^{-1}\bm{A}(\bm{C}+\lambda_g\bm{I}_{d_g})^{-1}\bm{b}.
\end{align*}

For the model selection, we can choose a model that minimizes $\widehat{J}(\widehat{f},\widehat{g})$ evaluated on a validation set.
However, as pointed out in \cite{yamane}, one cannot tell if $\widehat{J}(\widehat{f},\widehat{g})$ is small because $\widehat{f}$ is a good solution to the outer minimization, or $\widehat{g}$ is a poor solution to the inner maximization.
For this reason, $\widehat{\mu}_{\mathrm{dls}}(\bm{x}):=\widehat{\bm{\alpha}}_{\mathrm{dls}}^\top\bm{\phi}(\bm{x})$ can be unstable, and one cannot be confident about which model is the best.
The increased dimensionality of the search space because of the need to simultaneously select models for $f$ and $g$ can also make the model selection based on $\widehat{J}(\widehat{f},\widehat{g})$ challenging.

\section{Proposed Method}\label{sec:DPWLS}
We propose a novel estimator that enables simpler model selection than the DLS estimation while maintaining the essence of direct estimation as much as possible.
We avoid the minimax formulation of the objective functional as in (\ref{eq:minimax}) by estimating $\mu$ as a solution to the weighted least squares problem derived from the original problem (\ref{eq:obj_original}).
It can be constructed based on the sample averages from the separately observed samples like the DLS estimator, except we need to estimate $\pi(\bm{X})$ as a weight.
We term the proposed estimator as \emph{the directly weighted least squares (DWLS) estimator} since the pre-estimated PSD directly appears without inversion in the objective unlike the IPW estimators \cite{wooldridge2002,Wooldridge2007-vn,Seaman2013-mw}.

Consider the following weighted least squares problem:
\begin{align}
\min_{f\in L^2}E\left[\frac{w(\bm{X})}{\pi(\bm{X})}H_f(\bm{X})^2\right]=:Q_0(f|w),\label{eq:obj_wls}
\end{align}
where $w(\bm{X})$ is some weight depending on $\bm{X}$.
Rewriting the objective yields
\begin{align}
Q_0(f|w)=E[Tw(\bm{X})f(\bm{X})^2]-2E\left[Uw(\bm{X})f(\bm{X})\right]+E[S],\label{eq:obj_wls2}
\end{align}
where $S:=\frac{w(\bm{X})}{\pi(\bm{X})}\nu(\bm{X})^2$ is a constant and can therefore be safely ignored in the optimization.
Choosing $w=\pi$ clearly reduces the problem (\ref{eq:obj_wls}) to (\ref{eq:obj_original}).
Therefore, we can plug-in the pre-estimated $\pi(\bm{X})$ as a weight in practice to find a minimizer of the problem (\ref{eq:obj_original}).
However, $\pi(\bm{X})$ may not be the proper weight since it takes a negative value when $P(Z^{(1)}=1|\bm{X})<P(Z^{(0)}=1|\bm{X})$.
We can estimate $Q(f|\widehat{\pi}):=Q_0(f|\widehat{\pi})-E[S]$ based on the sample averages using the separately observed samples as
\begin{align*}
\widehat{Q}(f|\widehat{\pi})=\frac{1}{n_t}\sum_{i=1}^{n_t}r_{ti}t_i\widehat{\pi}(\bm{x}_{ti})f(\bm{x}_{ti})^2-\frac{2}{n_u}\sum_{i=1}^{n_u}r_{ui}u_i\widehat{\pi}(\bm{x}_{ui})f(\bm{x}_{ui}).
\end{align*}
Using the linear-in-parameter model for $f$ and the $\ell_2$-regularizer as in the previous section, we have
\begin{align*}
\widehat{Q}(f|\widehat{\pi})+\Omega(f)=\bm{\alpha}^\top\widetilde{\bm{A}}\bm{\alpha}-2\bm{\alpha}^\top\widetilde{\bm{b}}+\lambda_f\bm{\alpha}^\top\bm{\alpha},
\end{align*}
where
\begin{align*}
\widetilde{\bm{A}}:=\frac{1}{n_t}\sum_{i=1}^{n_t}r_{ti}t_i\widehat{\pi}(\bm{x}_{ti})\bm{\phi}(\bm{x}_{ti})\bm{\phi}(\bm{x}_{ti})^\top,\hspace{5mm}\widetilde{\bm{b}}:=\frac{1}{n_u}\sum_{i=1}^{n_u}r_{ui}u_i\widehat{\pi}(\bm{x}_{ui})\bm{\phi}(\bm{x}_{ui}).
\end{align*}
The analytical solution to $\min_{f\in F}\widehat{Q}(f|\widehat{\pi})+\Omega(f)$ can be obtained as $\widehat{\mu}_{\mathrm{dpw}}(\bm{x})=\widehat{\bm{\alpha}}_{\mathrm{dpw}}^\top\bm{\phi}(\bm{x})$, where
\begin{align*}
\widehat{\bm{\alpha}}_{\mathrm{dpw}}=\left(\widetilde{\bm{A}}+\lambda_f\bm{I}_{q_f}\right)^{-1}\widetilde{\bm{b}}.
\end{align*}

Although the DWLS objective is no longer the MSE, $\widehat{Q}(\widehat{f}|\widehat{\pi})$ is still a sufficient measure for model selection because $\widehat{\mu}_{\mathrm{dwls}}$ minimizing this objective is also a good estimator in terms of the true MSE as long as $\widehat{\pi}$ is sufficiently accurate.
Since the DWLS involves only a single minimization, the model selection is easier than in the DLS estimation.

\paragraph{Remark on the objective formulation}
We can consider the following least squares problem instead of (\ref{eq:obj_wls}):
\begin{align}
\min_{f\in L^2}E\left[\frac{\pi(\bm{X})}{w(\bm{X})}\left(f(\bm{X})-\mu(\bm{X})\right)^2\right]=:Q'_0(f|w).\label{eq:obj_ipw}
\end{align}
This objective can also be evaluated without the conditional mean function $\nu$ through a transformation similar to (\ref{eq:obj_wls2}):
\begin{align*}
Q'_0(f|w)=E\left[\frac{Tf(\bm{X})^2}{w(\bm{X})}\right]-2E\left[\frac{Uf(\bm{X})}{w(\bm{X})}\right]+E[S'],
\end{align*}
where $S':=\frac{\pi(\bm{X})}{w(\bm{X})}\mu(\bm{X})^2$.

We refer to the estimator based on this objective as \emph{the inverse weighted least squares (IWLS) estimator}.
This estimator tends to be imprecise when the PSD is extremely close to zero, which is a common issue among the IPW estimators \cite{wooldridge2002,Wooldridge2007-vn,Seaman2013-mw}.
Although the performance of such estimators can be improved by trimming small probabilities, determining the optimal threshold is nontrivial \cite{lee2011}.

Selecting the threshold is more complicated in the case of the IWLS as the PSD can be negative.
At points where the PSD is close to zero, the absolute value of the weights increases while the sign errors are more likely to occur.
As a result, a small estimation error in the PSD tends to have a large impact on the IWLS.
On the other hand, the impact of the estimation error of the PSD on the DWLS estimator is limited since the absolute value of the weights is small when a sign error occurs in the PSD estimation.
The weight in the proposed method is confined to $[-0.5,0.5]$, whereas the weight in the IWLS is not bounded at all.

Although the unweighted subsampling method has been recently studied to circumvents weighting samples with the inverse probability \cite{wang2020less}, it cannot be directly applied in this case because it requires the denominator to be a sampling probability.

\section{Performance Evaluation}
We test the performance of the proposed method with synthetic and real-world datasets.
The details of the datasets and other setups can be found in Appendix D.
We denote the separate estimation as the SEP.

\subsection{Synthetic Experiment}
\begin{figure*}[t]
  \def\@captype{table}
  \begin{minipage}[b]{.64\textwidth}
    \begin{center}
    \makeatletter
    \def\@captype{table}
    \makeatother
    \footnotesize
    \begin{tabular}{ccc|cccc}\hline
    Shape & $n$ & $q_x$ & DWLS & IWLS & SEP & DLS\\\hline
    &  & 1 & 0.65 $\pm$ 1.54 & 5.04 $\pm$ 16.5 & 1.80 $\pm$ 2.38 & \textbf{0.15} $\pm$ \textbf{0.20} \\
    & 10K & 5 & \textbf{0.40} $\pm$ \textbf{0.91} & 1.30 $\pm$ 3.21 & 3.46 $\pm$ 2.71 & 0.93 $\pm$ 1.17 \\
    \multirow{2}{*}{$h_{con}$} &  & 10 & \textbf{0.65} $\pm$ \textbf{1.00} & 4.96 $\pm$ 14.4 & 4.18 $\pm$ 3.81 & 1.81 $\pm$ 2.82 \\\cline{2-7}
    &  & 1 & \textbf{0.22} $\pm$ \textbf{0.46} & 53.5 $\pm$ 372 & 1.04 $\pm$ 0.93 & \textbf{0.08} $\pm$ \textbf{0.13} \\
    & 50K & 5 & \textbf{0.06} $\pm$ \textbf{0.07} & 0.35 $\pm$ 0.88 & 2.13 $\pm$ 2.00 & 0.41 $\pm$ 0.67 \\
    &  & 10 & \textbf{0.10} $\pm$ \textbf{0.11} & 1.22 $\pm$ 6.07 & 3.52 $\pm$ 1.92 & 0.41 $\pm$ 0.52 \\\hline
    &  & 1 & \textbf{0.09} $\pm$ \textbf{0.15} & 0.51 $\pm$ 1.25 & 0.25 $\pm$ 0.30 & 0.39 $\pm$ 0.30 \\
    & 10K & 5 & \textbf{0.14} $\pm$ \textbf{0.08} & 1.32 $\pm$ 6.72 & 0.58 $\pm$ 0.41 & 0.88 $\pm$ 0.52 \\
    \multirow{2}{*}{$h_{lin}$} &  & 10 & \textbf{0.61} $\pm$ \textbf{0.27} & 4.73 $\pm$ 22.4 & 1.39 $\pm$ 0.65 & 1.53 $\pm$ 0.75 \\\cline{2-7}
    &  & 1 & \textbf{0.02} $\pm$ \textbf{0.03} & 6.79 $\pm$ 53.1 & 0.11 $\pm$ 0.11 & 0.36 $\pm$ 0.33 \\
    & 50K & 5 & \textbf{0.04} $\pm$ \textbf{0.03} & 0.12 $\pm$ 0.16 & 0.36 $\pm$ 0.24 & 1.04 $\pm$ 0.53 \\
    &  & 10 &\textbf{ 0.32} $\pm$ \textbf{0.10} & 11.9 $\pm$ 112 & 0.76 $\pm$ 0.26 & 1.61 $\pm$ 0.89 \\\hline
    &  & 1 & \textbf{0.19} $\pm$ \textbf{0.26} & 0.72 $\pm$ 1.82 & 0.30 $\pm$ 0.31 & 0.32 $\pm$ 0.29 \\
    & 10K & 5 & \textbf{0.29} $\pm$ \textbf{0.23} & 12.1 $\pm$ 113 & 0.58 $\pm$ 0.40 & 0.97 $\pm$ 0.56 \\
    \multirow{2}{*}{$h_{log}$} &  & 10 & \textbf{0.72} $\pm$ \textbf{0.31} & 2.76 $\pm$ 7.65 & 1.11 $\pm$ 0.49 & 1.40 $\pm$ 0.72 \\\cline{2-7}
    &  & 1 & \textbf{0.06} $\pm$ \textbf{0.08} & 0.30 $\pm$ 0.88 & 0.22 $\pm$ 0.16 & 0.32 $\pm$ 0.30 \\
    & 50K & 5 & \textbf{0.10} $\pm$ \textbf{0.04} & 0.18 $\pm$ 0.13 & 0.56 $\pm$ 0.31 & 0.75 $\pm$ 0.62 \\
    &  & 10 & \textbf{0.30} $\pm$ \textbf{0.10} & 55.3 $\pm$ 550 & 0.86 $\pm$ 0.28 & 1.20 $\pm$ 0.99 \\\hline
    \end{tabular}
    \caption{The mean and standard deviation of the MSE over 100 trials with $\gamma=0$.
    The results are multiplied by 100 (constant), 10 (linear) and 10 (logistic), respectively.
    The bold face denotes the best and comparative results according to the two-sided Wilcoxon signed-rank test at the significance level of 5\%.}
    \label{tab:synth}
    \end{center}
  \end{minipage}
  \hfill
  \begin{minipage}[b]{.34\textwidth}
  \begin{center}
    \includegraphics[width=\textwidth]{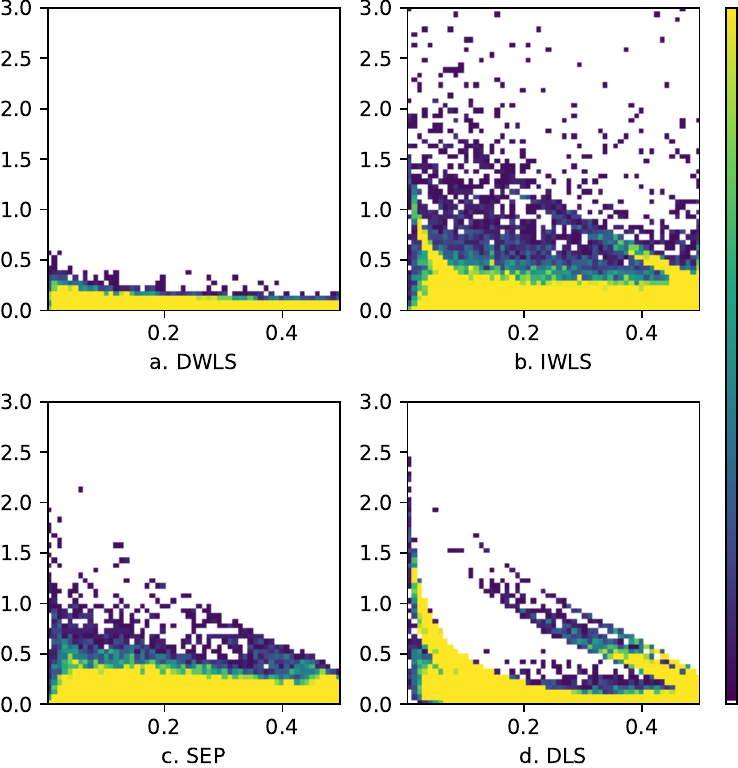}
    \caption{Relation between the squared error (y-axis) and the PSD (x-axis) when $\mu(\bm{X})$ is linear, $q_x=5$ and $n=10000$.
    Each point is coloured according to the spatial density.}
    \label{fig:synth}
  \end{center}
  \end{minipage}
\end{figure*}

We generated synthetic data with the different shapes of $\mu$, the covariates' dimension $q_x$, and the size of training samples.
The mean and standard deviation of the MSE over 100 trials are summarized in Table \ref{tab:synth}.
We conducted the Wilcoxon signed-rank test since the IWLS was highly unstable, which made it difficult to use the $t$-test.

Although we did not trim the PSD for the DWLS estimator, its performance was sufficiently stable to outperform the others in almost all the cases.
The DWLS had larger errors than the DLS only when $\mu$ was constant and $q_x=1$.
This may be because the benefit of direct estimation outweighed the difficulties of hyperparameter tuning of the DLS since the one-dimensional constant $\mu$ is the easiest to learn.
This result supports the effectiveness of our approach: avoiding the minimax formulation but without the inverse PSD.

On the other hand, the IWLS estimator often worked terribly poorly even with the weight trimming.
The IWLS might be affected by the small PSD most severely since it uses the inverse PSD both in the estimation and hyperparameter tuning.
The performance of the DLS was not as poor, but it had a larger MSE than SEP in many cases, indicating that hyperparameter tuning did not work as well as the DWLS.

Figure \ref{fig:synth} shows the relation between the squared error of each estimator and the PSD for the linear $\mu(\bm{X})$ when $q_x=5$ and $n=10000$.
The area with a dense plot is colored in light.
The squared error of the DWLS shown in Figure \ref{fig:synth}a is kept very small except it is slightly larger when the PSD is close to zero.
This result demonstrates the robustness of our method against the near-zero PSD, whereas the performance of the other estimators is affected more severely by the small PSD.
We can observe the instability of the IWLS again in Figure \ref{fig:synth}b, which shows the sporadic high errors over the entire PSD.
The squared error of the SEP displays the similar pattern.
In Figure \ref{fig:synth}d, the light-colored area is at a high position, indicating that the hyperparameter tuning of the DLS does not work well.

\subsection{Real-Data Analysis}
We used the dataset of the National Job Training Partnership Act (JTPA) study\footnote{\raggedright{https://www.upjohn.org/data-tools/employment-research-data-center/national-jtpa-study}}.
This is one of the largest RCT dataset for job training evaluations in the US with approximately 20000 participants, and it has been used in the several previous studies on causal inference \cite{bloom1997benefits,Abadie2002-ha,Donald2014-ca}.
We used the doubly robust (DR) estimator \cite{ogburn2015} trained with all joint samples as the pseudo true LATE since we do not know the true treatment effect for a real-world dataset.

The results are summarized in Table \ref{tab:jtpa}.
We report the mean, max, and min of $\widehat{\mu}(\bm{x})$ in addition to the root mean squared error (RMSE) to see the behavior of each estimator.
The IWLS had the smallest RMSE, but the difference from the DWLS was not statistically significant.
The mean, max, and min of the DWLS and IWLS also indicate they have similar performance.
The stability of the IWLS in contrast to the synthetic experiment is because there were no samples with a small PSD.
The relatively good performance of the SEP should be because of the same reason.
The performance of the DLS was highly unstable, reflecting the difficulty of the model selection based on the DLS objective.

\section{Conclusion}
We proposed a novel problem setting for LATE estimation by data combination.
Our setting is plausible enough to apply to many real-world problems.
The leading examples include O2O marketing and when there is non-random attrition in panel data.
We developed a practically useful method for estimating LATE from the separately observed datasets.
Our method overcomes the issue of the existing method in model selection by avoiding the minimax formulation of the objective.
Furthermore, our method can also avoid the common problem of the IPW methods since the PSD directly appears in our estimator without inversion.
Our method displayed the promising performance in the experiments on the synthetic and real-world datasets.
However, there are sometimes concerns on the homogeneity of the populations which the separate samples come from.
Extending the proposed method to account for heterogeneous samples and time-varying potential variables is an important future direction to further increase its usefulness in practice.

\begin{table*}[t]
    \centering
    \small
    \begin{tabular}{c|c|cccc}\hline
    & DR & DWLS & IWLS & SEP & DLS\\\hline
    RMSE & - & \textbf{252} $\pm$ \textbf{141} & \textbf{222} $\pm$ \textbf{145} & 280 $\pm$ 119 & 946 $\pm$ 663 \\
    Mean & 1778 & 1528 & 1559 & 1538 & 937 \\
    Max & 1837 & 1838 & 1924 & 2200 & 7966 \\
    Min & 1648 & 1013 & 977 & 919 & -1814 \\\hline
    \end{tabular}
    \caption{The results for performance evaluation using the JTPA dataset.
    The bold face denotes the best and comparative results according to the two-sided Wilcoxon signed-rank test at the significance level of 5\%.}
    \label{tab:jtpa}
\end{table*}

\bibliographystyle{abbrv}
\bibliography{late}

\newpage
\appendix
\section{Estimation of $\mu(\bm{X})$ when $P(\bm{X}^{(1)})\neq P(\bm{X}^{(0)})\neq P(\bm{X})$}
Consider weakening Assumption 1.1 to its conditional version:
\begin{align}
P(Y_1^{(1)},Y_0^{(1)},D_1^{(1)},D_0^{(1)}|\bm{X}^{(1)})=P(Y_1^{(0)},Y_0^{(0)},D_1^{(0)},D_0^{(0)}|\bm{X}^{(0)})=P(Y_1,Y_0,D_1,D_0|\bm{X}).\label{eq:invar}
\end{align}
Under this assumption, $\mu(\bm{X})$ is identified as follows:
\begin{align*}
\mu(\bm{x})=\frac{E[Y^{(1)}|\bm{X}^{(1)}=\bm{x}]-E[Y^{(0)}|\bm{X}^{(0)}=\bm{x}]}{E[D^{(1)}|\bm{X}^{(1)}=\bm{x}]-E[D^{(0)}|\bm{X}^{(0)}=\bm{x}]}.
\end{align*}
In this case, directly applying the DLS and DWLS may result in large bias since $P(\bm{X}^{(1)})\neq P(\bm{X}^{(0)})\neq P(\bm{X})$ in general.
We need density ratios to adjust a distribution over which the expectation is taken to apply the DLS and DWLS.
Unfortunately, we cannot estimate the density ratio $\frac{P(\bm{X}|D^{(k)}=1)}{P(\bm{X}^{(k)}|D^{(k)}=1)}$ when we have $\{\bm{x}_{di}^{(k)}\}_{i=1}^{n_d^{(k)}}$ and $\widehat{p}_d^{(k)}$ as in the data collection scheme introduced in Section 2.3.
We do not have any information on $P(\bm{X}|D^{(k)}=1)$ in this case.

If we can collect $\{(d_i^{(k)},\bm{x}_{di}^{(k)})\}_{i=1}^{n_d^{(k)}}\overset{\mathrm{iid}}{\sim}P(D^{(k)},\bm{X}^{(k)})$ instead of the pair $\{\bm{x}_{di}^{(k)}\}_{i=1}^{n_d^{(k)}}$ and $\widehat{p}_d^{(k)}$, we can modify the combined datasets explained in Section 4.1 to
\begin{align*}
\{(t_i,\bm{x}_{ti},r_{ti})\}_{i=1}^{n_t}&:=\left\{\left((-1)^{1-k}d_i^{(k)},\bm{x}_{di}^{(k)},\frac{n_d^{(1)}+n_d^{(0)}}{2n_d^{(k)}}\widehat{\rho}_k(\bm{x}_{di}^{(k)})\right)\right\}_{i=1,k=0}^{n_d^{(k)},1},\\
\{(u_i,\bm{x}_{ui},r_{ui})\}_{i=1}^{n_u}&:=\left\{\left((-1)^{1-k}y_i^{(k)},\bm{x}_i^{(k)},\frac{n^{(1)}+n^{(0)}}{2n^{(k)}}\widehat{\rho}_k(\bm{x}_i^{(k)})\right)\right\}_{i=1,k=0}^{n^{(k)},1},
\end{align*}
where $\rho_k(\bm{X}):=\frac{P(\bm{X})}{P(\bm{X}^{(k)})}$ for $k=0,1$, and estimate $\mu(\bm{X})$ with the DLS and DWLS using the above datasets.
Various methods can be employed to estimate density ratios (Nguyen, Wainwright, and Jordan 2010; Sugiyama, Suzuki, and Kanamori 2012; Liu et al. 2017; Kato and Teshima 2021).

However, it is no longer direct estimation in the original sense because we need to estimate the density ratios.
Although we have to maintain Assumption 1.1 to make direct estimation possible, this limitation does not detract significantly from the value of our work since our setting is plausible enough to apply to many real-world problems even with Assumption 1.1 as explained in the main text.
Note that the concern for the time-varying potential variables when using RCS data does not vanish even if we adopt the condition (\ref{eq:invar}) since the conditional distribution may also change over time.

\paragraph{Relation to the Standard Setting}
Consider the estimation of $\mu(\bm{X})$ using separate samples drawn from $P(Y,Z,\bm{X})$ and $P(D,Z,\bm{X})$ under the standard assumptions (Abadie 2003; Fr\"{o}lich 2007; Tan 2006; Ogburn, Rotnitzky, and Robins 2015).
Recalling that $\mu(\bm{X})$ can be identified as
\begin{align*}
\mu(\bm{X})=\frac{E[Y|Z=1,\bm{X}]-E[Y|Z=0,\bm{X}]}{E[D|Z=1,\bm{X}]-E[D|Z=0,\bm{X}]},
\end{align*}
we can see that it is essentially the same problem as our setting with the condition (\ref{eq:invar}).
We need $\frac{P(\bm{X})}{P(\bm{X}|Z=z)}$ for $z=0,1$ in this case.

\section{Proof of Theorem 1}
\begin{proof}
By Assumption 1.1, 2.1 and 2.2, we have
\begin{align}
E[D^{(1)}-D^{(0)}|\bm{X}]&=E[Z^{(1)}D_1+(1-Z^{(1)})D_0|\bm{X}]-E[Z^{(0)}D_1+(1-Z^{(0)})D_0|\bm{X}]\notag\\
&=E[(Z^{(1)}-Z^{(0)})(D_1-D_0)|\bm{X}]\notag\\
&=E[Z^{(1)}-Z^{(0)}|\bm{X}]E[D_1-D_0|\bm{X}].\label{eq:ddif}
\end{align}
Similarly, we can show
\begin{align}
E[Y^{(1)}-Y^{(0)}|\bm{X}]&=E[D^{(1)}Y_1+(1-D^{(1)})Y_0|\bm{X}]-E[D^{(0)}Y_1+(1-D^{(0)})Y_0|\bm{X}]\notag\\
&=E[(D^{(1)}-D^{(0)})(Y_1-Y_0)|\bm{X}]\notag\\
&=E[(Z^{(1)}-Z^{(0)})(D_1-D_0)(Y_1-Y_0)|\bm{X}]\notag\\
&=E[Z^{(1)}-Z^{(0)}|\bm{X}]E[(D_1-D_0)(Y_1-Y_0)|\bm{X}]\notag\\
&=E[Z^{(1)}-Z^{(0)}|\bm{X}]P(D_1-D_0=1|\bm{X})E[Y_1-Y_0|D_1-D_0=1\bm{X}].\label{eq:ydif}
\end{align}
It holds that $E[D_1-D_0|\bm{X}]=P(D_1-D_0=1|\bm{X})$ since $D_1-D_0\in\{0,1\}$ by Assumption 2.3.
Moreover, $D_1-D_0=1$ implies $D_1>D_0$ since $D_1$ and $D_0$ are binary.
Note that $E[Z^{(1)}-Z^{(0)}|\bm{X}]\neq0$ and $E[D_1-D_0|\bm{X}]\neq0$ by Assumption 1.2 and 2.4, respectively.
Combining these facts with (\ref{eq:ddif}) and (\ref{eq:ydif}) yields
\begin{align*}
\frac{E[Y^{(1)}-Y^{(0)}|\bm{X}]}{E[D^{(1)}-D^{(0)}|\bm{X}]}&=\frac{E[Z^{(1)}-Z^{(0)}|\bm{X}]E[D_1-D_0|\bm{X}]E[Y_1-Y_0|D_1-D_0=1\bm{X}]}{E[Z^{(1)}-Z^{(0)}|\bm{X}]E[D_1-D_0|\bm{X}]}\\
&=E[Y_1-Y_0|D_1-D_0=1,\bm{X}]\\
&=E[Y_1-Y_0|D_1>D_0,\bm{X}]\\
&=\mu(\bm{X}).
\end{align*}
\end{proof}

\section{Direct Estimation of PSD}
We cannot employ ordinary regression methods on $\{(t_i,\bm{x}_{ti},r_{ti})\}_{i=1}^{n_t}$ to obtain an estimator of $\pi(\bm{X})$ since it does not follow $P(T|\bm{X})$ even when $n_d^{(1)}=n_d^{(0)}$.
We have to develop an estimator which incorporates the covariates of $\{(u_i,\bm{x}_{ui},r_{ui})\}_{i=1}^{n_u}$.
Moreover, an estimator of $\pi(\bm{X})$ should satisfy the constraint $\pi(\bm{X})\in[-0.5,0.5]$.
In the following, we develop a direct estimation method which is basically an application of the least squares method for estimating the density ratio (Kanamori, Hido, and Sugiyama 2009; Sugiyama, Suzuki, and Kanamori 2012).

In order to develop an estimator satisfying the constraint, consider the least squares estimation of $\pi(\bm{X})+0.5\in[0,1]$ by minimizing the following squared error:
\begin{align*}
E[(f(\bm{X})-\pi(\bm{X})-0.5)^2]&=E[f(\bm{X})^2]-2E[Tf(\bm{X})]-E[f(\bm{X})]+E[(\pi(\bm{X})+0.5)^2].
\end{align*}
The last term can be safely ignored in optimization.
We can approximate the first three terms as
\begin{align*}
\frac{1}{n_u}\sum_{i=1}^{n_u}r_{ui}f(\bm{x}_{ui})^2-\frac{2}{n_t}\sum_{i=1}^{n_t}r_{ti}t_if(\bm{x}_{ti})-\frac{1}{n_u}\sum_{i=1}^{n_u}r_{ui}f(\bm{x}_{ui}).
\end{align*}
Using the linear-in-parameter model and $\ell_2$-regularizer, we have an analytical solution $\widetilde{\bm{\alpha}}_{0.5+}^\top\bm{\phi}(\bm{x})$, where
\begin{align*}
\widetilde{\bm{\alpha}}_{0.5+}=\left(\frac{1}{n_u}\sum_{i=1}^{n_u}r_{ui}\bm{\phi}(\bm{x}_{ui})\bm{\phi}(\bm{x}_{ui})^\top+\lambda_f\bm{I}_{q_f}\right)^{-1}\left(\frac{1}{n_t}\sum_{i=1}^{n_t}r_{ti}t_i\bm{\phi}(\bm{x}_{ti})+\frac{1}{2n_u}\sum_{i=1}^{n_u}r_{ui}\bm{\phi}(\bm{x}_{ui})\right).
\end{align*}
When we use a non-negative basis function such as the Gaussian kernel, we can modify $\widetilde{\bm{\alpha}}_{0.5+}$ as $\widehat{\bm{\alpha}}_{0.5+}:=\max\{\widetilde{\bm{\alpha}}_{0.5+},\bm{0}_{q_f}\}$ to force $\widehat{\bm{\alpha}}_{0.5+}^\top\bm{\phi}(\bm{x})$ to be non-negative as well.
Here, the $\max$ operator extracts larger values element-wise.
Similarly, we can construct an estimator of $-\pi(\bm{X})+0.5\in[0,1]$ as $\widehat{\bm{\alpha}}_{0.5-}^\top\bm{\phi}(\bm{x})$, where $\widehat{\bm{\alpha}}_{0.5-}:=\max\{\widetilde{\bm{\alpha}}_{0.5-},\bm{0}_{q_f}\}$ and
\begin{align*}
\widetilde{\bm{\alpha}}_{0.5-}=\left(\frac{1}{n_u}\sum_{i=1}^{n_u}r_{ui}\bm{\phi}(\bm{x}_{ui})\bm{\phi}(\bm{x}_{ui})^\top+\lambda_f\bm{I}_{q_f}\right)^{-1}\left(-\frac{1}{n_t}\sum_{i=1}^{n_t}r_{ti}t_i\bm{\phi}(\bm{x}_{ti})+\frac{1}{2n_u}\sum_{i=1}^{n_u}r_{ui}\bm{\phi}(\bm{x}_{ui})\right).
\end{align*}
Finally, we normalize an estimate of $\pi(\bm{X})+0.5$ at a given point $\bm{x}$ as
\begin{align*}
\widehat{(\pi+0.5)}(\bm{x})=\frac{\widehat{\bm{\alpha}}_{0.5+}^\top\bm{\phi}(\bm{x})}{(\widehat{\bm{\alpha}}_{0.5+}^\top+\widehat{\bm{\alpha}}_{0.5-}^\top)\bm{\phi}(\bm{x})},
\end{align*}
since $(\pi(\bm{X})+0.5)+(-\pi(\bm{X})+0.5)=1$.
We can ensure $\widehat{(\pi+0.5)}(\bm{x})\leq1$ for any $\bm{x}$ by this normalization.
Therefore, we can obtain an estimator that respects the constraint by setting $\widehat{\pi}(\bm{x}):=\widehat{(\pi+0.5)}(\bm{x})-0.5$.

\paragraph{Estimation in 1E2RD}
We can further tighten the restriction to $\pi(\bm{X})\in[0,0.5]$ in the 1E2RD since the PSD is non-negative by Assumption 2.3 when $P(Z^{(0)}=1|\bm{X})=0$.
In this case, we can obtain the non-negative estimator of $\pi(\bm{X})$ as $\widehat{\bm{\alpha}}_{\pi}^\top\bm{\phi}(\bm{x})$, where $\widehat{\bm{\alpha}}_{\pi}:=\max\{\widetilde{\bm{\alpha}}_{\pi},\bm{0}_{q_f}\}$ and
\begin{align*}
\widetilde{\bm{\alpha}}_\pi=\left(\frac{1}{n_u}\sum_{i=1}^{n_u}r_{ui}\bm{\phi}(\bm{x}_{ui})\bm{\phi}(\bm{x}_{ui})^\top+\lambda_f\bm{I}_{q_f}\right)^{-1}\left(\frac{1}{n_t}\sum_{i=1}^{n_t}r_{ti}t_i\bm{\phi}(\bm{x}_{ti})\right).
\end{align*}
Then, we can normalize the estimator so that it takes values no larger than 0.5 as
\begin{align*}
\widehat{\pi}(\bm{x})=\frac{\widehat{\bm{\alpha}}_{\pi}^\top\bm{\phi}(\bm{x})}{2(\widehat{\bm{\alpha}}_{\pi}^\top+\widehat{\bm{\alpha}}_{0.5-}^\top)\bm{\phi}(\bm{x})},
\end{align*}
since $\pi(\bm{X})+(-\pi(\bm{X})+0.5)=0.5$.

\section{Experimental Setups}

\subsection{Synthetic Experiment}
The data generating process used for the experiments is as follows:
\begin{align*}
Y'_0=\varsigma(\bm{1}_{q_x}^\top&\bm{X})+(0.2D_1+0.1D_0)\times\bm{1}_{q_x}^\top\bm{X},\hspace{3mm}Y'_1=Y'_0+h(\bm{X},D_1,D_0),\\
&Y_0=Y'_0+\varepsilon_0,\hspace{3mm}Y_1=Y'_1+\varepsilon_1,\hspace{3mm}\bm{X}\sim N(\bm{0}_{q_x},\bm{\Sigma}_{q_x}),\\
P(&Z^{(1)}=1|\bm{X})=\varsigma(1+0.2\times\bm{1}_{q_x}^\top\bm{X}),\hspace{3mm}P(Z^{(0)}=1|\bm{X})=0,\\
P(D_1&=1|\bm{X})=\varsigma(\gamma+4+\bm{1}_{q_x}^\top\bm{X}),\hspace{2mm}P(D_0=1|\bm{X})=\varsigma(\gamma+\bm{1}_{q_x}^\top\bm{X}),
\end{align*}
where $\varsigma(v)=(1+e^{-v})^{-1}$, $\bm{1}_{q_x}$ is a $q_x\times1$ vector with all elements equal to 1, $\bm{0}_{q_x}$ is a $q_x\times1$ zero vector and $\bm{\Sigma}_{q_x}$ is a $q_x\times q_x$ positive semidefinite matrix with diagonal elements equal to 1 and the absolute value of nondiagonal elements less than 0.5.
$\varepsilon_0$ and $\varepsilon_1$ are mean zero random noise with $\mathrm{var}(\varepsilon_d)=0.5$ for $d=0,1$ and $\mathrm{cov}(\varepsilon_0,\varepsilon_1)=0.2$.
$\gamma$ is an experimental parameter for controlling the ratio of the extreme PSD, and we set $\gamma=0$ for the experiment in the main text.
The experiments with $\gamma=-1,1$ yielded similar findings, and their results can be found in Appendix E.

We implemented the experiments with three different shapes of the treatment effect function $h$:
\begin{align*}
h_{\mathrm{con}}(\bm{X},D_1,D_0)&=0.2+0.3D_1+0.1D_0,\\
h_{\mathrm{lin}}(\bm{X},D_1,D_0)&=(0.1+0.15D_1+0.05D_0)\times\bm{1}_{q_x}^\top\bm{X},\\
h_{\mathrm{log}}(\bm{X},D_1,D_0)&=\varsigma((1+0.2D_1+0.1D_0)\times\bm{1}_{q_x}^\top\bm{X}).
\end{align*}
Note that $\mu(\bm{X})=h(\bm{X},1,0)$.
We separately generated four sets of the training samples $\{\bm{x}_{di}^{(k)}\}_{i=1}^{n_d^{(k)}}$, $\{(y_i^{(k)},\bm{x}_i^{(k)})\}_{i=1}^{n^{(k)}}$ with $n_d^{(k)}=n^{(k)}=10000$ or $50000$ for $k=0,1$, validation sets with the same sample size as the training samples and a test set with 10000 samples.

We compared the proposed method (DWLS) with the IWLS, the separate estimation (SEP) and the DLS.
The Gaussian kernels centered at 100 randomly chosen training samples were used for the basis functions $\bm{\phi}$ and $\bm{\psi}$.
We also used the same kernel for the kernel ridge regression to estimate $\nu$.
The estimation of $\pi$ was performed using the direct estimation method explained in Appendix C.
The value of $\widehat{\pi}$ was trimmed to $0.15$ to prevent the weight from being too large for the SEP and IWLS.
This trimming was not performed for the DWLS.
The bandwidth $h$ of the Gaussian kernel and the regularization parameter $\lambda$ were selected based on the model selection criterion corresponding to each estimator on validation sets.
We used optuna (Akiba et al. 2019) for the hyperparameter tuning with 100 searches, and we set the search space for $h$ and $\lambda$ to $[1,10]$ and $[10^{-5},10^5]$, respectively.

\subsection{Real Data Analysis}
We treat the JTPA dataset as if it was collected from the 1E2RD under one-sided noncompliance since the noncompliance structure was almost one-sided in the JTPA study.
Only 2\% of the participants in the control group received the job training treatment, whereas approximately 40\% in the treatment group did not receive the treatment.
We excluded the participants with $Z=0$ and $D=1$ from our analysis to focus on the one-sided noncompliance structure.
We used the samples with $Z=0,1$ as $K=1$ and the samples with $Z=0$ as $K=0$.
Note that, therefore, this sampling procedure is with replacement.
Assumption 1.1 is satisfied with this sampling since $Z$ was assigned completely at random in the JTPA study.

The outcome of interest here is the total earnings over 30 months after a random assignment.
The covariates used in our analysis included gender, age, earnings in the previous year, a dummy for the black, a dummy for Hispanic, a dummy for high school graduates, a dummy for the married and a dummy for whether a participant worked for at least 12 weeks in the 12 months preceding the random assignment.
We excluded the samples with earnings in the previous year larger than 90 percentile to stabilize the estimation.
We applied the Yeo-Johnson transformation (Yeo and Johnson 2000) to the age and earnings in the previous year since they are non-negative and thus have a highly skewed distribution.
All variables were normalized to have zero mean and unit standard deviation.
We evaluated the performance of each estimator based on the root mean squared error computed using a 100-fold cross-validation.
For hyperparameter tuning, we calculated the model selection criterion corresponding to each estimator by 10-fold cross-validation within the training sets.
Weight trimming was not performed since $\widehat{\pi}(\bm{X})$ was far from zero for all $\bm{X}$.

\section{Additional Experimental Results}

\subsection{Relation between the Squared Error and the PSD}
We present the graphs of the relation between the squared error and the PSD for all the cases.
Similar to Figure 2, these figures show that the performance deterioration of the DWLS at points with the small PSD is limited compared to the other estimators.

\subsection{Sensitivity Analysis}
In addition to the synthetic experiment in the main text, we conducted the experiments with $\gamma=-1,1$ to investigate the change in the performance of each estimator according to the ratio of the small PSD.
Figure \ref{fig:psdhist} shows the histograms of the PSD with different values of $\gamma$.
It is clear that the proportion of the small PSD increases as $\gamma$ increases.

Table \ref{tab:gamma1} and \ref{tab:gamma-1} presents the results with $\gamma=1,-1$, respectively.
In both cases, we obtained the results similar to the case with $\gamma=0$: the DWLS had the smallest MSE for all the cases except the constant $\mu$ with $q_x=1$.
When $\gamma=1$, the relative advantage of the DWLS over the other estimators decreased for all the cases.
Since the DWLS directly uses the PSD, the efficiency of the DWLS estimator declines as the proportion of the small PSD increases.
In this case, the performance of the DWLS could be improved by the weight trimming so that samples with the near-zero PSD have at least a certain impact on estimation.
When $\gamma=-1$, the MSE decreased overall.
The variance of the IWLS got smaller than in the case with the other value of $\gamma$, but it is still unstable compared to the other methods.

\begin{figure}
\begin{tabular}{ccc}
    \begin{minipage}[t]{0.3\textwidth}
    \begin{center}
        \includegraphics[width=\textwidth]{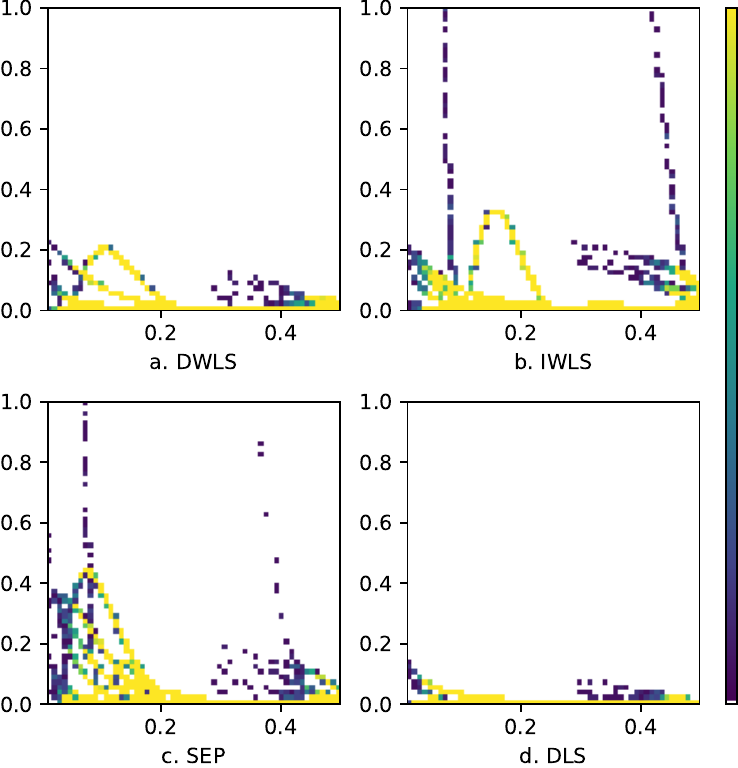}
        \caption{Constant, $q_x=1$, $n=10000$.}
    \end{center}
    \end{minipage}&
    \begin{minipage}[t]{0.3\textwidth}
    \begin{center}
        \includegraphics[width=\textwidth]{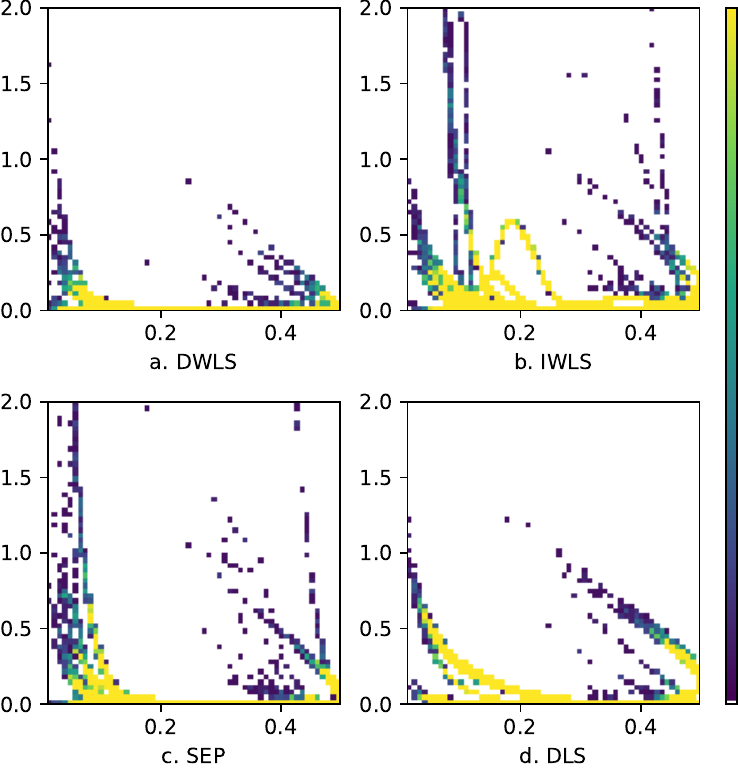}
        \caption{Linear, $q_x=1$, $n=10000$.}
    \end{center}
    \end{minipage}&
    \begin{minipage}[t]{0.3\textwidth}
    \begin{center}
        \includegraphics[width=\textwidth]{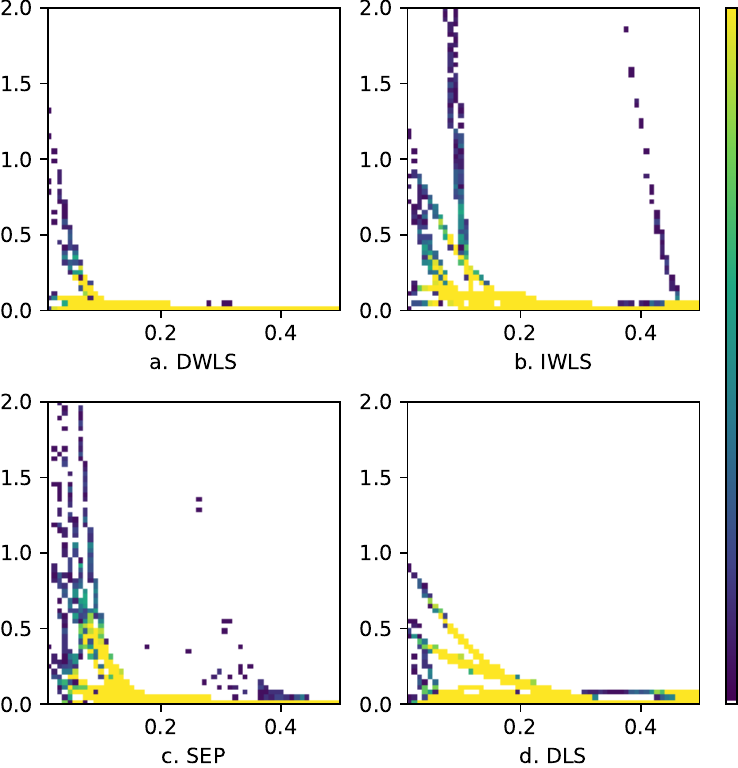}
        \caption{Logistic, $q_x=1$, $n=10000$.}
    \end{center}
    \end{minipage}\\
    \begin{minipage}[t]{0.3\textwidth}
    \begin{center}
        \includegraphics[width=\textwidth]{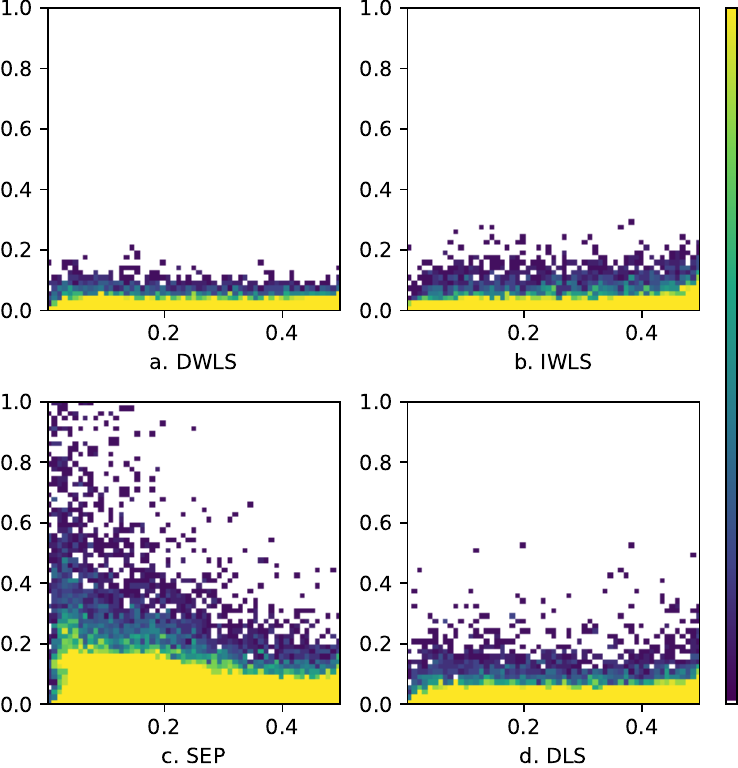}
        \caption{Constant, $q_x=5$, $n=10000$.}
    \end{center}
    \end{minipage}&
    \begin{minipage}[t]{0.3\textwidth}
    \begin{center}
        \includegraphics[width=\textwidth]{figs/linear5_10000_1.0_0.0.pdf}
        \caption{Linear, $q_x=5$, $n=10000$.}
    \end{center}
    \end{minipage}&
    \begin{minipage}[t]{0.3\textwidth}
    \begin{center}
        \includegraphics[width=\textwidth]{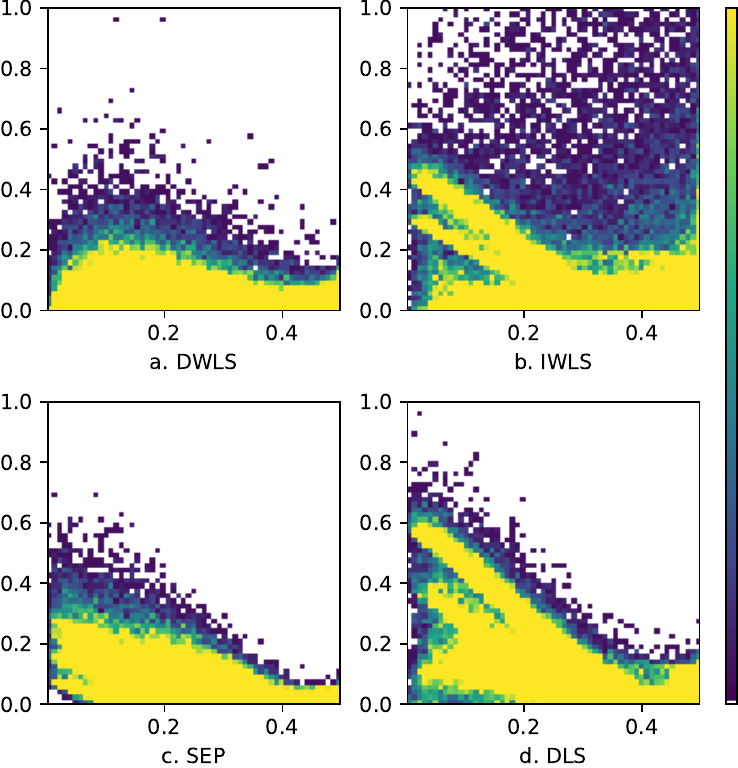}
        \caption{Logistic, $q_x=5$, $n=10000$.}
    \end{center}
    \end{minipage}\\
    \begin{minipage}[t]{0.3\textwidth}
    \begin{center}
        \includegraphics[width=\textwidth]{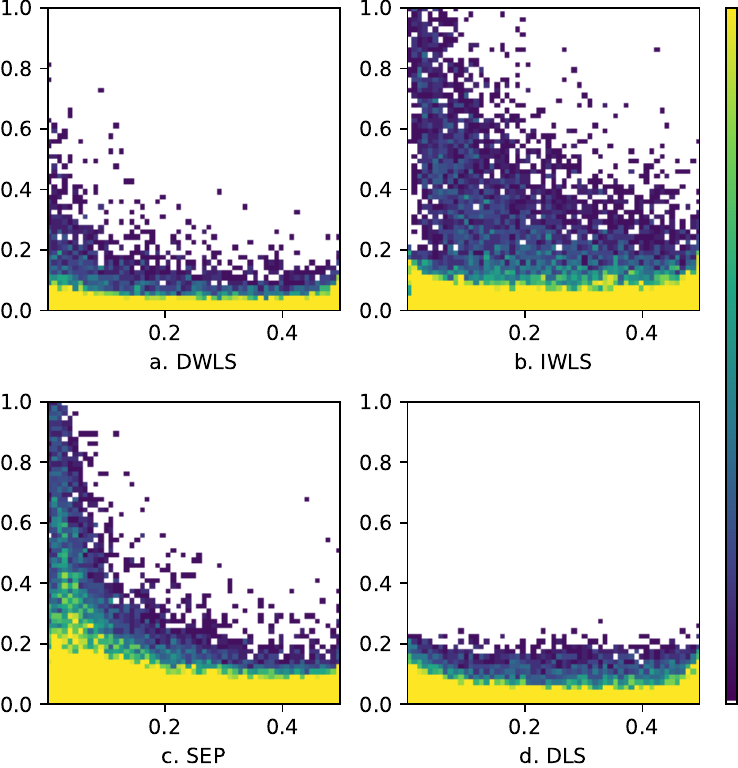}
        \caption{Constant, $q_x=10$, $n=10000$.}
    \end{center}
    \end{minipage}&
    \begin{minipage}[t]{0.3\textwidth}
    \begin{center}
        \includegraphics[width=\textwidth]{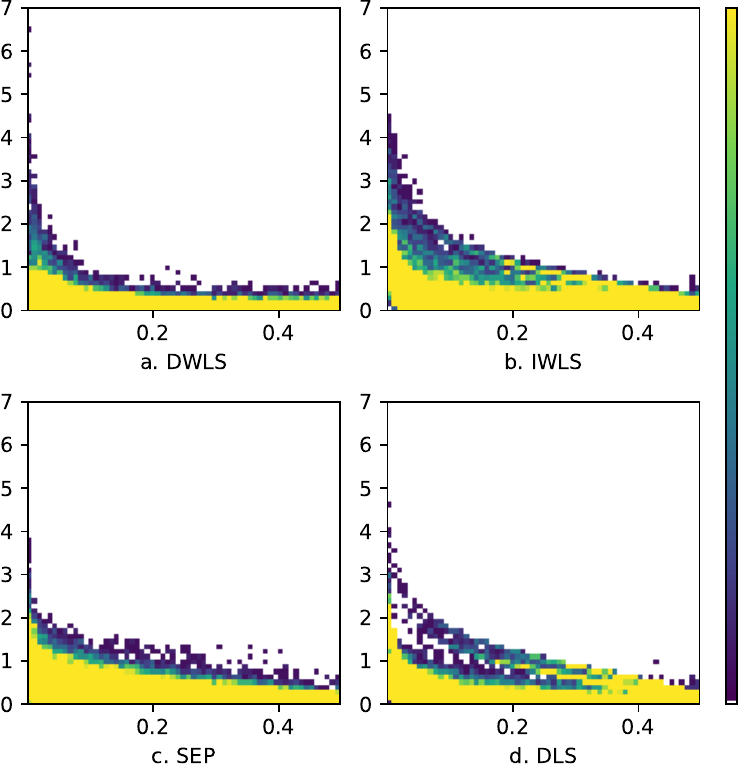}
        \caption{Linear, $q_x=10$, $n=10000$.}
    \end{center}
    \end{minipage}&
    \begin{minipage}[t]{0.3\textwidth}
    \begin{center}
        \includegraphics[width=\textwidth]{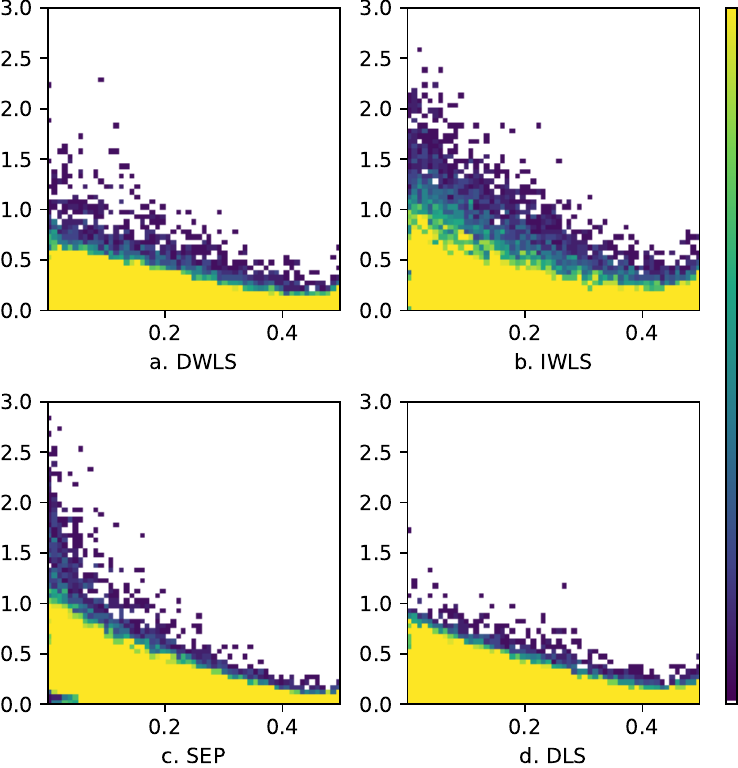}
        \caption{Logistic, $q_x=10$, $n=10000$.}
    \end{center}
    \end{minipage}
\end{tabular}
\end{figure}

\begin{figure}
\begin{tabular}{ccc}
    \begin{minipage}[t]{0.3\textwidth}
    \begin{center}
        \includegraphics[width=\textwidth]{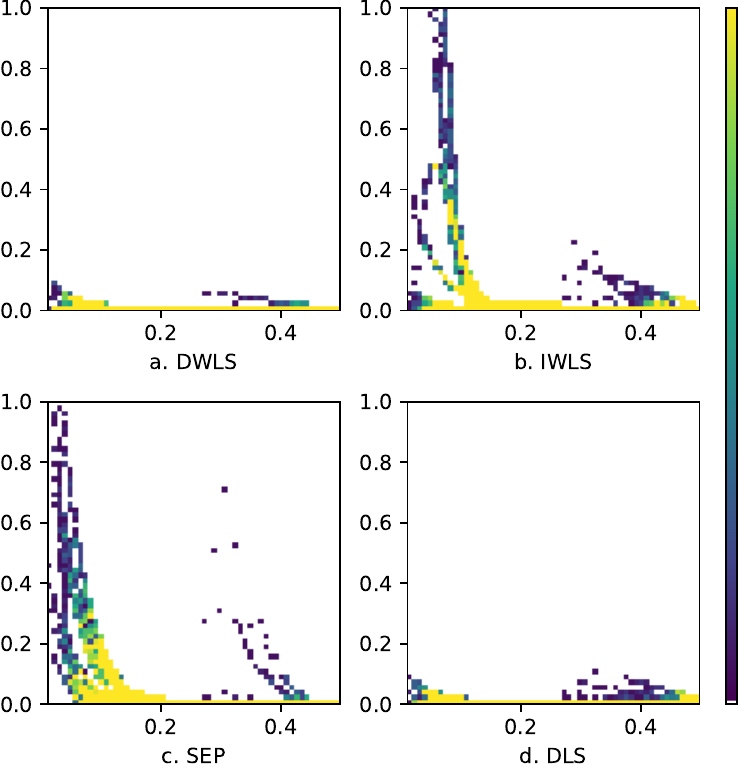}
        \caption{Constant, $q_x=1$, $n=50000$.}
    \end{center}
    \end{minipage}&
    \begin{minipage}[t]{0.3\textwidth}
    \begin{center}
        \includegraphics[width=\textwidth]{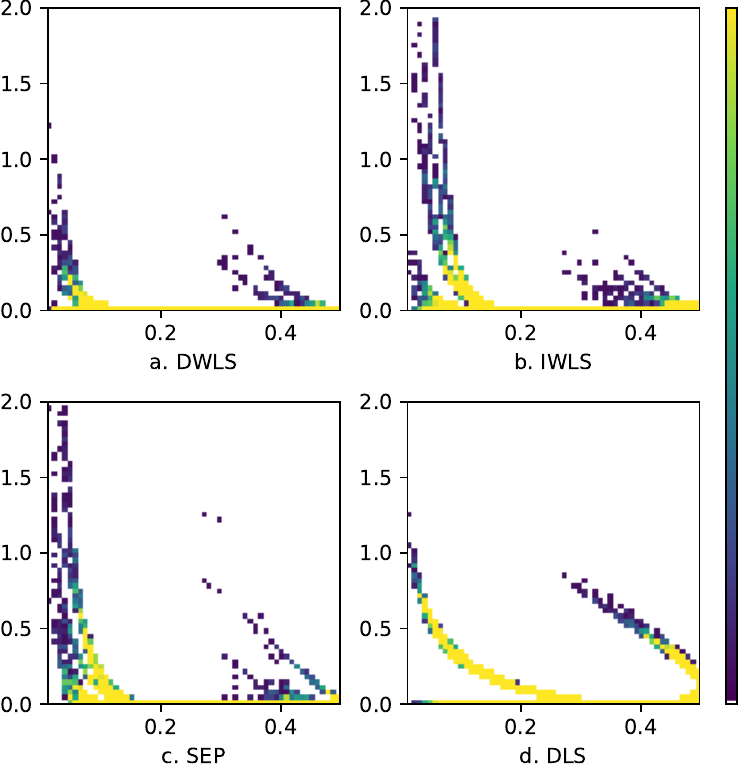}
        \caption{Linear, $q_x=1$, $n=50000$.}
    \end{center}
    \end{minipage}&
    \begin{minipage}[t]{0.3\textwidth}
    \begin{center}
        \includegraphics[width=\textwidth]{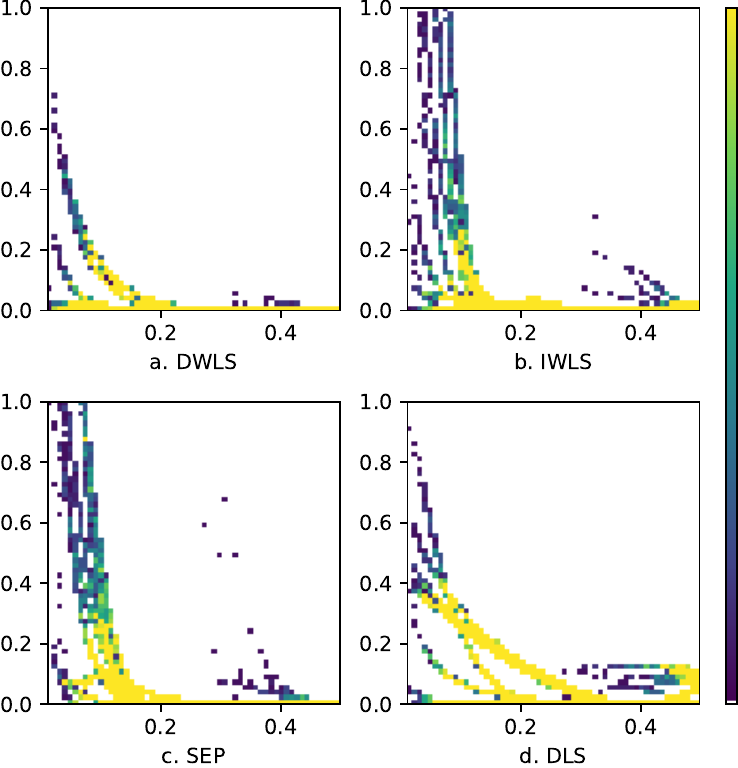}
        \caption{Logistic, $q_x=1$, $n=50000$.}
    \end{center}
    \end{minipage}\\
    \begin{minipage}[t]{0.3\textwidth}
    \begin{center}
        \includegraphics[width=\textwidth]{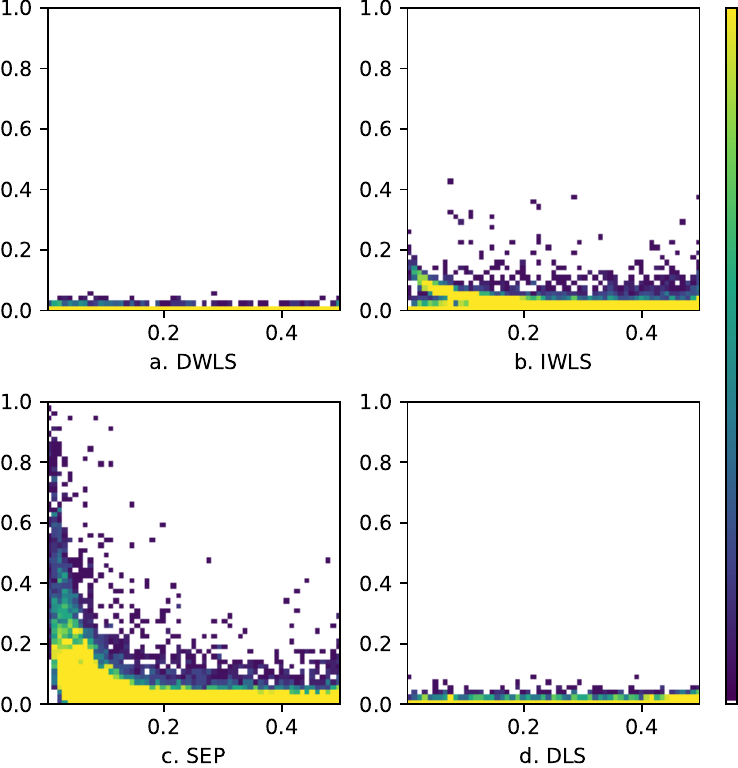}
        \caption{Constant, $q_x=5$, $n=50000$.}
    \end{center}
    \end{minipage}&
    \begin{minipage}[t]{0.3\textwidth}
    \begin{center}
        \includegraphics[width=\textwidth]{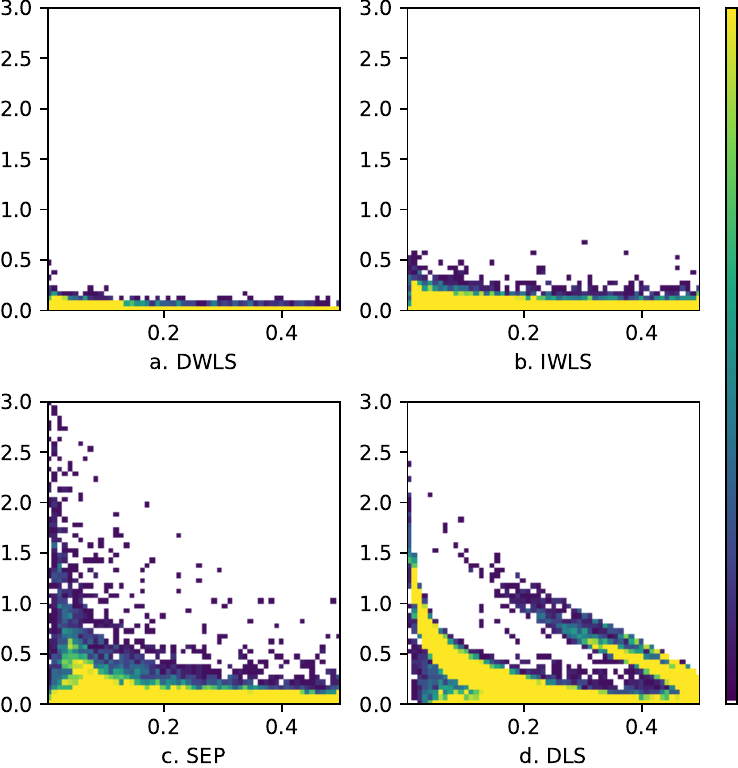}
        \caption{Linear, $q_x=5$, $n=50000$.}
    \end{center}
    \end{minipage}&
    \begin{minipage}[t]{0.3\textwidth}
    \begin{center}
        \includegraphics[width=\textwidth]{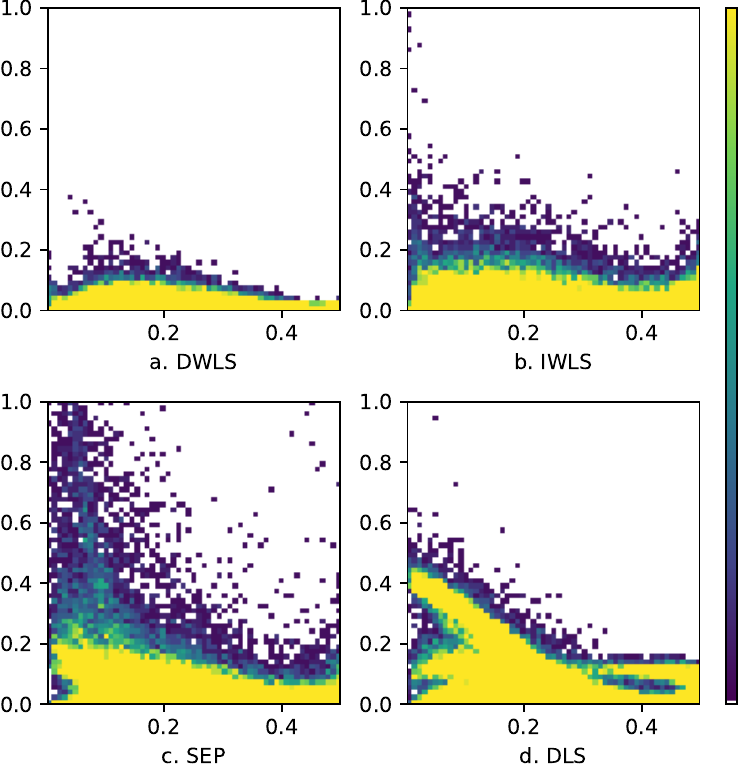}
        \caption{Logistic, $q_x=5$, $n=50000$.}
    \end{center}
    \end{minipage}\\
    \begin{minipage}[t]{0.3\textwidth}
    \begin{center}
        \includegraphics[width=\textwidth]{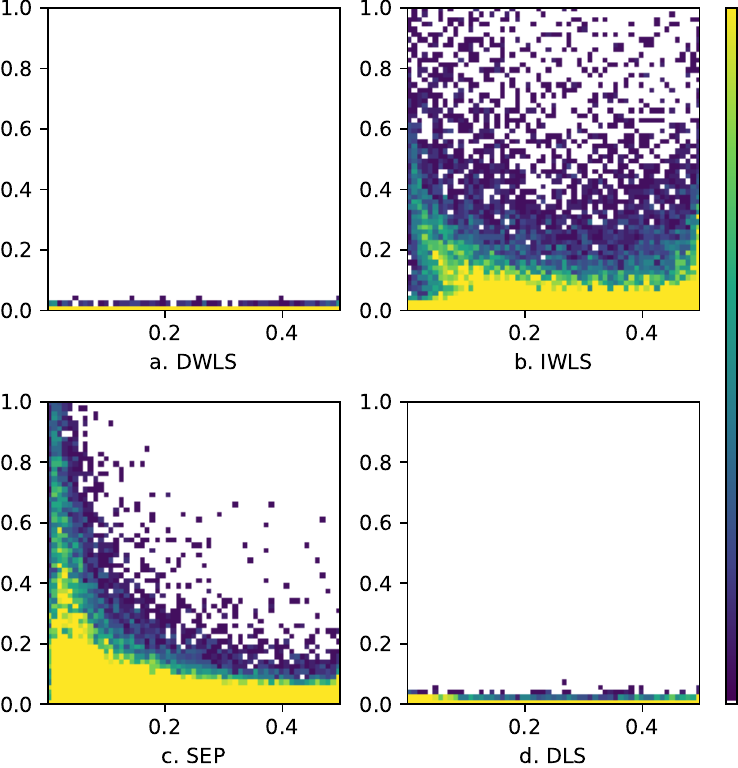}
        \caption{Constant, $q_x=10$, $n=50000$.}
    \end{center}
    \end{minipage}&
    \begin{minipage}[t]{0.3\textwidth}
    \begin{center}
        \includegraphics[width=\textwidth]{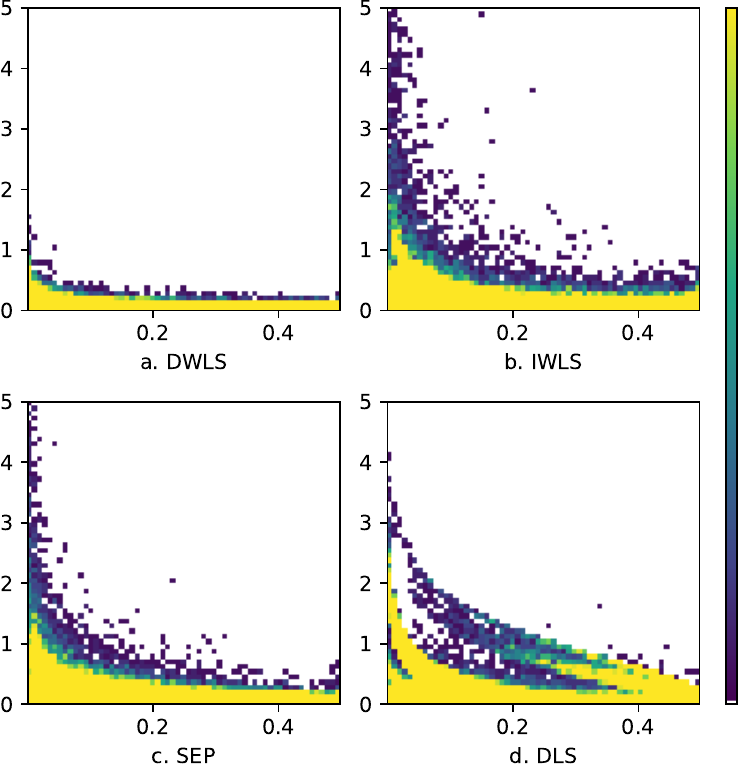}
        \caption{Linear, $q_x=10$, $n=50000$.}
    \end{center}
    \end{minipage}&
    \begin{minipage}[t]{0.3\textwidth}
    \begin{center}
        \includegraphics[width=\textwidth]{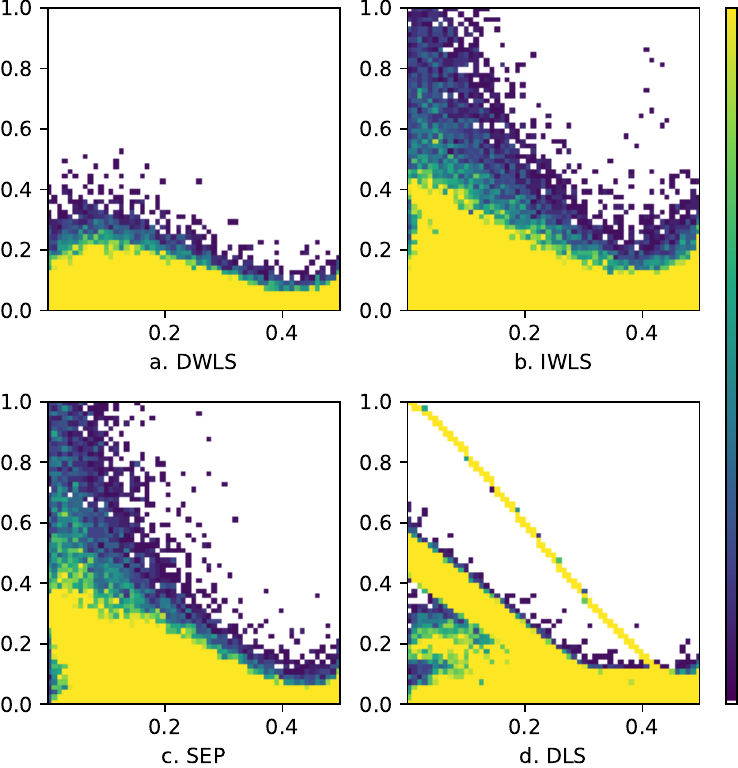}
        \caption{Logistic, $q_x=10$, $n=50000$.}
    \end{center}
    \end{minipage}
\end{tabular}
\end{figure}
\newpage

\begin{figure}[t]
    \begin{minipage}[b]{0.33\textwidth}
    \begin{center}
        \includegraphics[width=\textwidth]{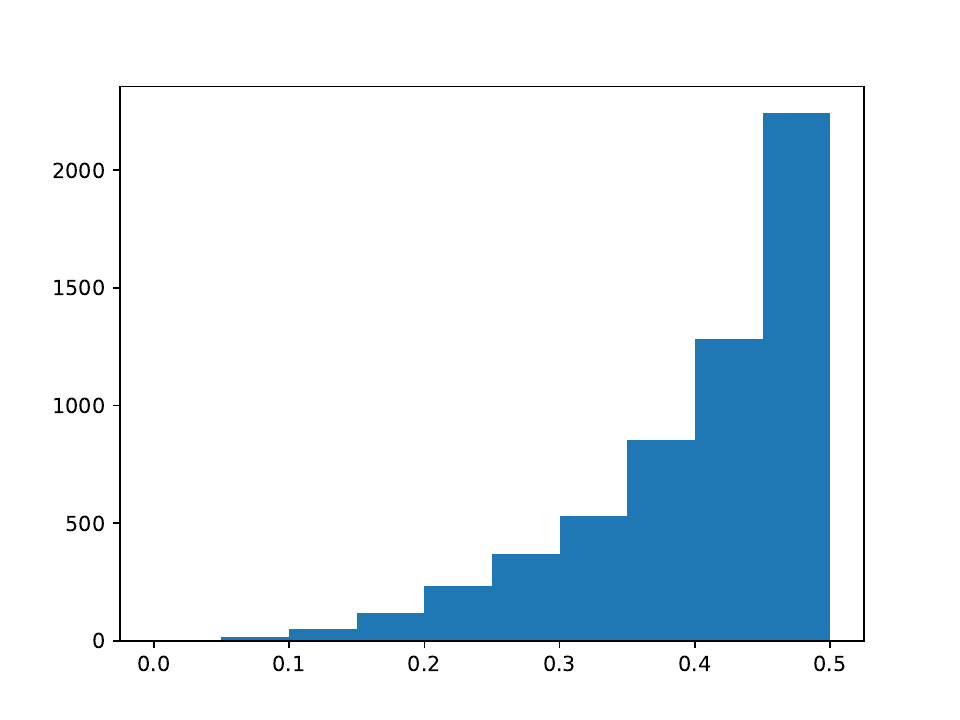}
        \subcaption{$\gamma=-1$}
    \end{center}
    \end{minipage}
    \begin{minipage}[b]{0.33\textwidth}
    \begin{center}
        \includegraphics[width=\textwidth]{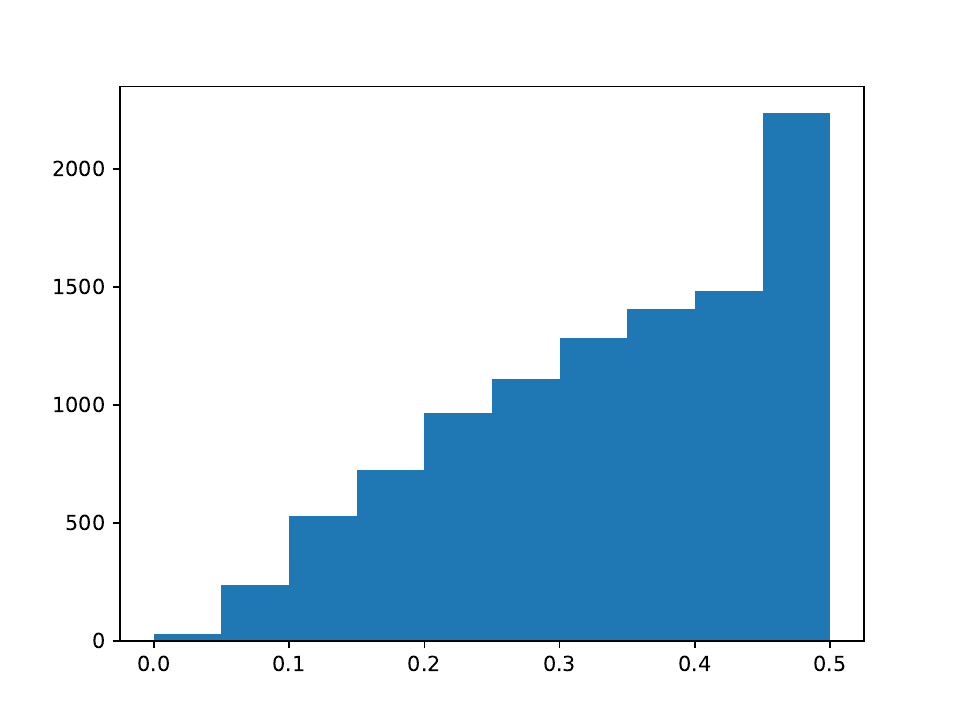}
        \subcaption{$\gamma=0$}
    \end{center}
    \end{minipage}
    \begin{minipage}[b]{0.33\textwidth}
    \begin{center}
        \includegraphics[width=\textwidth]{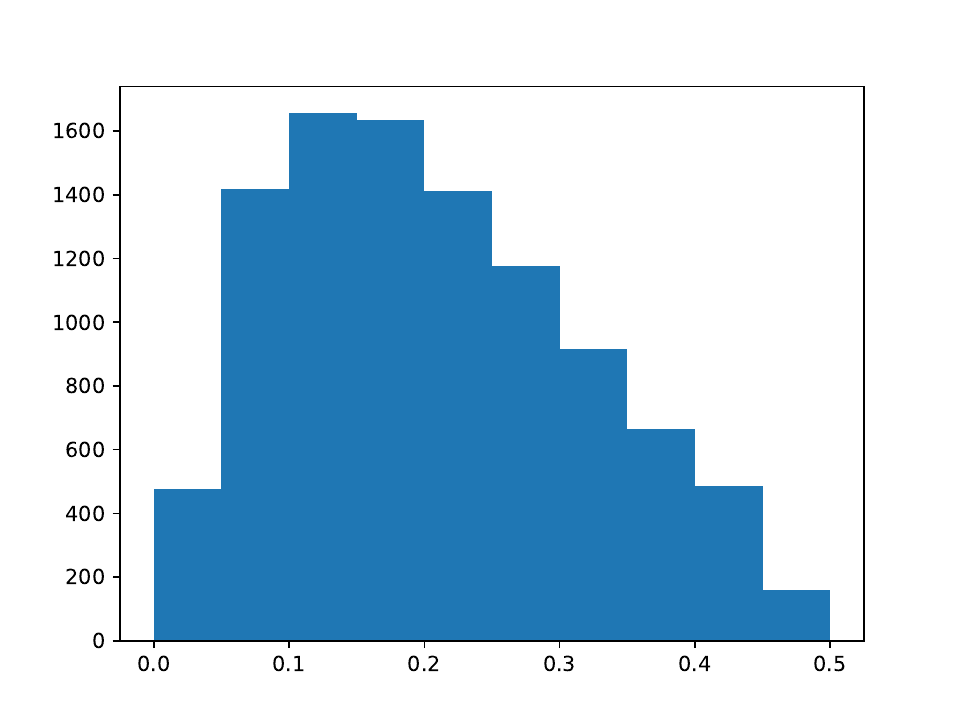}
        \subcaption{$\gamma=1$}
    \end{center}
    \end{minipage}
    \caption{Histograms of the PSD.}
    \label{fig:psdhist}
\end{figure}

\newpage
\begin{table}[t]
    \centering
    \footnotesize
    \begin{tabular}{ccc|cccc}\hline
    Shape & $n$ & $q_x$ & DWLS & IWLS & SEP & DLS\\\hline
    &  & 1 & \textbf{0.11} $\pm$ \textbf{0.32} & 404 $\pm$ 3482 & 0.55 $\pm$ 0.23 & \textbf{0.10} $\pm$ \textbf{0.46} \\
    & 10K & 5 & \textbf{0.10} $\pm$ \textbf{0.23} & 0.79 $\pm$ 5.09 & 0.70 $\pm$ 0.32 & 0.23 $\pm$ 0.30 \\
    \multirow{2}{*}{$h_{\mathrm{con}}$} &  & 10 & \textbf{0.09} $\pm$ \textbf{0.19} & 30.5 $\pm$ 225 & 0.71 $\pm$ 0.36 & 0.28 $\pm$ 0.30 \\\cline{2-7}
    &  & 1 & \textbf{0.05} $\pm$ \textbf{0.14} & 85.4 $\pm$ 623 & 0.56 $\pm$ 0.11 & \textbf{0.02} $\pm$ \textbf{0.04} \\
    & 50K & 5 & \textbf{0.01} $\pm$ \textbf{0.02} & 0.07 $\pm$ 0.17 & 0.63 $\pm$ 0.13 & 0.05 $\pm$ 0.10 \\
    &  & 10 & \textbf{0.02} $\pm$ \textbf{0.02} & 158 $\pm$ 1551 & 0.66 $\pm$ 0.19 & 0.11 $\pm$ 0.27 \\\hline
    &  & 1 & \textbf{0.02} $\pm$ \textbf{0.08} & 578 $\pm$ 5723 & 0.03 $\pm$ 0.02 & 0.04 $\pm$ 0.03 \\
    & 10K & 5 & \textbf{0.05} $\pm$ \textbf{0.05} & 0.24 $\pm$ 0.90 & 0.08 $\pm$ 0.04 & 0.14 $\pm$ 0.06 \\
    \multirow{2}{*}{$h_{\mathrm{lin}}$} &  & 10 & \textbf{0.15} $\pm$ \textbf{0.29} & 3.23 $\pm$ 19.8 & 0.15 $\pm$ 0.05 & 0.20 $\pm$ 0.09 \\\cline{2-7}
    &  & 1 & \textbf{0.01} $\pm$ \textbf{0.02} & 10.3 $\pm$ 50.2 & 0.02 $\pm$ 0.01 & 0.05 $\pm$ 0.04 \\
    & 50K & 5 & \textbf{0.01} $\pm$ \textbf{0.01} & 0.03 $\pm$ 0.04 & 0.06 $\pm$ 0.02 & 0.13 $\pm$ 0.06 \\
    &  & 10 & \textbf{0.05} $\pm$ \textbf{0.03} & 0.60 $\pm$ 2.94 & 0.10 $\pm$ 0.02 & 0.18 $\pm$ 0.10 \\\hline
    &  & 1 & \textbf{0.04} $\pm$ \textbf{0.08} & 68.4 $\pm$ 573 & 0.13 $\pm$ 0.03 & 0.05 $\pm$ 0.04 \\
    & 10K & 5 & \textbf{0.10} $\pm$ \textbf{0.08} & 0.52 $\pm$ 3.58 & 0.18 $\pm$ 0.04 & 0.20 $\pm$ 0.07 \\
    \multirow{2}{*}{$h_{\mathrm{log}}$} &  & 10 & \textbf{0.17} $\pm$ \textbf{0.18} & 9.18 $\pm$ 63.1 & 0.23 $\pm$ 0.04 & 0.25 $\pm$ 0.07 \\\cline{2-7}
    &  & 1 & \textbf{0.02} $\pm$ \textbf{0.03} & 17.2 $\pm$ 94.5 & 0.13 $\pm$ 0.02 & 0.05 $\pm$ 0.04 \\
    & 50K & 5 & \textbf{0.03} $\pm$ \textbf{0.02} & 0.05 $\pm$ 0.07 & 0.17 $\pm$ 0.02 & 0.12 $\pm$ 0.07 \\
    &  & 10 & \textbf{0.08} $\pm$ \textbf{0.03} & 0.32 $\pm$ 2.12 & 0.21 $\pm$ 0.03 & 0.17 $\pm$ 0.08 \\\hline
    \end{tabular}
    \caption{The mean and standard deviation of the MSE over 100 trials with $\gamma=1$.
    The results are multiplied by 10 (constant), 1 (linear) and 1 (logistic), respectively.
    The bold face denotes the best and comparative results according to the two-sided Wilcoxon signed-rank test at the significance level of 5\%.}
    \label{tab:gamma1}
\end{table}

\begin{table}[t]
    \centering
    \footnotesize
    \begin{tabular}{ccc|cccc}\hline
    Shape & $n$ & $q_x$ & DWLS & IWLS & SEP & DLS\\\hline
    &  & 1 & 0.30 $\pm$ 0.59 & 1.19 $\pm$ 3.72 & 0.76 $\pm$ 0.88 & \textbf{0.14} $\pm$ \textbf{0.30} \\
    & 10K & 5 & \textbf{0.24} $\pm$ \textbf{0.29} & 0.56 $\pm$ 1.02 & 1.97 $\pm$ 1.89 & 0.66 $\pm$ 1.12 \\
    \multirow{2}{*}{$h_{\mathrm{con}}$} &  & 10 & \textbf{0.43} $\pm$ \textbf{0.52} & 2280 $\pm$ 22666 & 2.18 $\pm$ 1.92 & 0.88 $\pm$ 1.08 \\\cline{2-7}
    &  & 1 & \textbf{0.11} $\pm$ \textbf{0.27} & 0.46 $\pm$ 1.33 & 0.40 $\pm$ 0.34 & \textbf{0.06} $\pm$ \textbf{0.11} \\
    & 50K & 5 & \textbf{0.05} $\pm$ \textbf{0.07} & 0.16 $\pm$ 0.28 & 1.74 $\pm$ 0.96 & 0.18 $\pm$ 0.26 \\
    &  & 10 & \textbf{0.09} $\pm$ \textbf{0.21} & 0.39 $\pm$ 1.13 & 2.06 $\pm$ 0.96 & 0.36 $\pm$ 0.46 \\\hline
    &  & 1 & \textbf{0.06} $\pm$ \textbf{0.17} & 0.19 $\pm$ 0.44 & 0.10 $\pm$ 0.11 & 0.40 $\pm$ 0.29 \\
    & 10K & 5 & \textbf{0.11} $\pm$ \textbf{0.09} & 0.26 $\pm$ 0.20 & 0.29 $\pm$ 0.17 & 0.75 $\pm$ 0.50 \\
    \multirow{2}{*}{$h_{\mathrm{lin}}$} &  & 10 & \textbf{0.50} $\pm$ \textbf{0.17} & 18.8 $\pm$ 178 & 1.02 $\pm$ 0.37 & 1.19 $\pm$ 0.70 \\\cline{2-7}
    &  & 1 & \textbf{0.01} $\pm$ \textbf{0.02} & 0.05 $\pm$ 0.15 & 0.05 $\pm$ 0.05 & 0.35 $\pm$ 0.31 \\
    & 50K & 5 & \textbf{0.03} $\pm$ \textbf{0.02} & 0.06 $\pm$ 0.06 & 0.18 $\pm$ 0.12 & 0.90 $\pm$ 0.52 \\
    &  & 10 & \textbf{0.25} $\pm$ \textbf{0.06} & \textbf{0.32} $\pm$ \textbf{0.24} & 0.63 $\pm$ 0.18 & 1.36 $\pm$ 0.86 \\\hline
    &  & 1 & \textbf{0.07} $\pm$ \textbf{0.08} & 0.26 $\pm$ 0.60 & 0.15 $\pm$ 0.13 & 0.28 $\pm$ 0.29 \\
    & 10K & 5 & \textbf{0.17} $\pm$ \textbf{0.12} & 0.35 $\pm$ 0.26 & 0.50 $\pm$ 0.28 & 0.56 $\pm$ 0.33 \\
    \multirow{2}{*}{$h_{\mathrm{log}}$} &  & 10 & \textbf{0.47} $\pm$ \textbf{0.18} & 321 $\pm$ 3187 & 0.87 $\pm$ 0.33 & 0.85 $\pm$ 0.60 \\\cline{2-7}
    &  & 1 & \textbf{0.03} $\pm$ \textbf{0.04} & 0.07 $\pm$ 0.16 & 0.09 $\pm$ 0.06 & 0.32 $\pm$ 0.40 \\
    & 50K & 5 & \textbf{0.06} $\pm$ \textbf{0.01} & 0.11 $\pm$ 0.06 & 0.35 $\pm$ 0.15 & 0.70 $\pm$ 0.45 \\
    &  & 10 & \textbf{0.21} $\pm$ \textbf{0.05} & 0.32 $\pm$ 0.14 & 0.57 $\pm$ 0.16 & 0.78 $\pm$ 0.77 \\\hline
    \end{tabular}
    \caption{The mean and standard deviation of the MSE over 100 trials with $\gamma=-1$.
    The results are multiplied by 100 (constant), 10 (linear) and 10 (logistic), respectively.
    The bold face denotes the best and comparative results according to the two-sided Wilcoxon signed-rank test at the significance level of 5\%.}
    \label{tab:gamma-1}
\end{table}

\end{document}